\documentclass{article}

\usepackage{PRIMEarxiv}

\usepackage[utf8]{inputenc}
\usepackage[T1]{fontenc}
\usepackage{booktabs}
\usepackage{amsfonts}
\usepackage{nicefrac}
\usepackage{microtype}
\usepackage{xcolor}
\usepackage{graphicx}
\usepackage{multirow}
\usepackage{makecell}
\graphicspath{{figures/}}
\usepackage{amsthm}
\newtheorem{remark}{Remark}
\newtheorem{assumption}{Assumption}

\usepackage{algorithm}
\usepackage{algpseudocode}
\algrenewcommand\algorithmicrequire{\textbf{Input:}}
\algrenewcommand\algorithmicensure{\textbf{Output:}}
\usepackage{wrapfig}
\usepackage{caption}
\usepackage{float}
\usepackage{placeins}
\usepackage{flafter}
\usepackage{amsmath}

\usepackage{mathtools}
\usepackage{subcaption}
\usepackage{tabularx}
\usepackage{wrapfig}
\usepackage{amssymb}
\usepackage{enumitem}
\usepackage{placeins}
\usepackage{url}
\usepackage{xurl}
\usepackage[numbers,sort&compress]{natbib}
\usepackage[hidelinks]{hyperref}
\usepackage[capitalize,noabbrev]{cleveref}

\newcolumntype{C}[1]{>{\centering\arraybackslash}p{#1}}

\newcommand{\answerTODO}[1][]{\textcolor{red}{\bfseries [TODO]}}
\newcommand{\justificationTODO}[1][]{\textcolor{red}{\bfseries [TODO]}}

\title{Not All Rollouts Are Worth Learning: On Trajectory Valuation for Post-Training Reinforcement Learning}

\author{%
  Xuesong Jia$^{*}$, Ziao Yang$^{*}$, Zhanhe Huang, Hongfu Liu \\
  Department of Computer Science, Brandeis University\\
  $^*$Co-First Author with Equal Contribution\\
  \texttt{\{jasjia,ziaoyang,erichzh,hongfuliu\}@brandeis.edu } \\
}

\begin{document}

\maketitle

\begin{abstract}
We consider the problem of trajectory valuation in reinforcement learning: how to identify and mitigate detrimental trajectories during online training. Unlike classification, where data valuation relies on fixed training and validation sets, reinforcement learning involves dynamically generated trajectories without explicit validation signals, making conventional influence-based methods inapplicable. We propose Dynamic Trajectory Valuation (DTV), a simple and efficient framework that estimates trajectory utility at the mini-batch level and filters detrimental trajectories based solely on gradient information. By operating at the optimization level, DTV integrates seamlessly with existing reinforcement learning pipelines with minimal overhead. Extensive experiments across diverse settings, including PPO, GRPO, and DPO, demonstrate that DTV consistently improves performance, enhances data efficiency, and stabilizes optimization.
\end{abstract}

\section{Introduction}
Reinforcement learning is a paradigm in which an agent learns sequential decision-making policies through interaction with an environment, with the goal of maximizing long-term cumulative rewards~\citep{sutton2018reinforcement}. Unlike classification with fixed labeled datasets, reinforcement learning data---commonly referred to as trajectories---is generated online through the agent's rollouts. As a result, the quality of collected trajectories directly affects learning performance, and not all trajectories contribute positively to policy optimization~\citep{schaul2015prioritized,shu2026learning}.

This naturally motivates a data-centric perspective in reinforcement learning, where the goal is to evaluate and curate trajectories to improve learning outcomes. Data-centric learning has recently emerged as a promising direction for improving model performance by assessing the contribution of individual samples to a target objective. A central task in this paradigm is data valuation, which aims to quantify the utility of each sample~\citep{koh2017understanding,kwon2023datainf,park2023trak}. A straightforward approach is to measure performance differences with and without a sample, but such retraining-based methods are computationally prohibitive. Influence functions provide an efficient alternative by estimating sample influence without retraining, but introduce new challenges in approximating the inverse Hessian. Prior work has developed various approximations, including stochastic recursive methods, random projections, Kronecker-factored eigendecomposition, and low-rank approximations, as well as Hessian-free variants for scalability~\citep{grosse2023studying,yang2024revisit}. While these methods have been extensively studied in classification, their extension to reinforcement learning remains limited.

Extending influence-based data valuation from classification to reinforcement learning is non-trivial. Classification relies on fixed training and validation sets, enabling valuation through validation performance, whereas reinforcement learning involves dynamically generated trajectories without a fixed validation set. Moreover, trajectory utility in reinforcement learning is inherently policy-dependent and shaped by temporal credit assignment and reward dynamics, making it difficult to define consistent impact measures. Finally, reinforcement learning requires dynamic, online valuation as training progresses, raising additional challenges in efficiency.

Recent work~\citep{hu2025snapshot,shu2026learning,li2026learnalign,yang2026gradalign} has taken initial steps toward data valuation and selection in reinforcement learning. Existing methods rely on replay buffers and surrogate impact functions, verifiable feedback, or trusted validation gradients to estimate data utility. However, these setting-specific designs raise concerns about their generalizability across different reinforcement learning paradigms.

\textbf{Contributions}. In this paper, we study trajectory valuation in reinforcement learning, aiming to prevent detrimental trajectories from model updates. We summarize our key contributions as follows:

\vspace{-2mm}
\begin{itemize}[wide=10pt, leftmargin=*]

\item We provide a principled formulation for trajectory valuation in general reinforcement learning settings, addressing the challenges of dynamic data generation, the absence of validation signals, and policy-dependent utility within a unified framework. This formulation establishes a unified perspective for trajectory valuation across diverse reinforcement learning pipelines.

\item We propose Dynamic Trajectory Valuation (DTV), a simple and efficient method that dynamically identifies and filters detrimental trajectories at the mini-batch level. Our method operates purely on gradients and integrates seamlessly with existing reinforcement learning pipelines.

\item We conduct extensive experiments to demonstrate the effectiveness of DTV across diverse reinforcement learning settings, including Proximal Policy Optimization (PPO), Group Relative Policy Optimization (GRPO), and Direct Preference Optimization (DPO). The results show consistent performance improvements, enhanced data efficiency, and more stable optimization, validating the benefits of dynamic trajectory valuation in practice.
\end{itemize}
% \vspace{-2mm}

\section{Related Work}
Here, we review prior work on data-centric learning in reinforcement learning and influence function-based methods, and position our approach in relation to existing literature.

\subsection{Data-Centric Learning in Reinforcement Learning} 
% \textbf{Experience selection and sample efficiency}.
A line of work focuses on improving sample efficiency by prioritizing or filtering training data~\citep{schaul2015prioritized,li2021revisiting,jiang2021prioritized,bengio2009curriculum}, most commonly at the level of individual transitions (i.e., experiences), as in prioritized experience replay, or via structured data scheduling such as curriculum learning. These approaches implicitly acknowledge that different data points contribute unequally to learning. However, they primarily operate on local signals---such as temporal-difference error or immediate rewards---to estimate importance, without explicitly modeling the contribution of entire trajectories. As a result, they may fail to capture long-range dependencies and the global effect of a trajectory on policy improvement.

Moving beyond transition-level selection, recent approaches in reinforcement learning from human feedback operate directly at the level of complete trajectories or response sequences~\citep{christiano2017deep,ouyang2022training}. Methods such as PPO~\citep{schulman2017proximal}, GRPO~\citep{shao2024deepseekmath}, and DPO~\citep{rafailov2023direct} treat each rollout as a training unit, where the quality of entire outputs determines the learning signal. In practice, data quality is often controlled via heuristic strategies such as rejection sampling or reward thresholding~\citep{bai2022training,nakano2021webgpt}. While effective in large-scale systems, these approaches still rely on coarse signals and lack a principled framework for evaluating trajectories based on their contribution to learning.

Recent work~\citep{hu2025snapshot,shu2026learning,li2026learnalign,yang2026gradalign} has begun to explore the adaptation of data valuation and selection in reinforcement learning\footnote{We provide an additional comparison of these methods in Appendix~\ref{app:relatedwork}.}. IIF~\citep{hu2025snapshot} estimates data influence using replay buffers and algorithm-specific surrogate objectives; LearnAlign~\citep{li2026learnalign} weights gradient alignment by learnability derived from verifiable ground-truth feedback; and GradAlign~\citep{yang2026gradalign} evaluates candidate updates against gradients from a trusted validation set. While this line of work represents a promising direction, existing methods often rely on auxiliary signals, which may limit their general applicability across different reinforcement learning paradigms.

Taken together, existing approaches either rely on local heuristics at the transition level, employ coarse filtering strategies, or approximate reinforcement learning through supervised surrogates. A unified and principled framework for trajectory-level data valuation remains lacking.

\subsection{Influence Function}
Influence functions, a mainstream tool for data valuation, were originally developed in robust statistics~\citep{hampel1974influence, cook1982residuals, martin1986influence} to quantify the sensitivity of model parameters to perturbations in training data. They were introduced to the machine learning community by~\citet{koh2017understanding} and have since been widely used to estimate sample influence with respect to validation performance.

To address the high computational cost of influence functions, several works propose efficient approximations of the inverse Hessian. LiSSA~\citep{koh2017understanding} leverages Hessian--vector products for stochastic approximation, while EKFAC~\citep{grosse2023studying} exploits structured eigenvalue decompositions. DataInf~\citep{kwon2023datainf} further simplifies influence estimation by leveraging a rank-1 structure of the empirical Hessian, and TRAK~\citep{park2023trak} employs random projections to obtain a tractable kernel-based approximation. More recently, Hessian-free influence methods have been proposed, which approximate the inverse Hessian with an identity matrix for scalability~\citep{charpiat2019input, pruthi2020estimating, yang2024revisit, killamsetty2021grad}.

Beyond classical influence formulations, several recent works revisit influence estimation from an optimization-aware and non-convex perspective. SGD-Influence~\citep{hara2019data} tracks the stochastic gradient descent trajectory and estimates the effect of removing a training sample by reusing intermediate iterates, thereby relaxing the local convexity assumption. MoSo~\citep{tan2023data} models data pruning by approximating a moving-one-sample-out update along the optimization path, while Z0-Inf~\citep{kokhlikyan2025z0} develops a zeroth-order estimator that avoids explicit gradients and Hessian computations.

Beyond static estimation, dynamic data valuation has recently attracted increasing attention~\citep{wang2024data,yang2025layer}. Early approaches approximate dynamic influence by aggregating influence estimates across training checkpoints~\citep{pruthi2020estimating,grosse2023studying,park2023trak}. However, such aggregation fails to capture the evolving nature of influence and may cancel out conflicting signals. More recent work directly estimates data value at each training step, enabling truly dynamic valuation~\citep{wang2024data,yang2025layer}.

In this paper, rather than adapting influence functions from classification settings, we generalize them to reinforcement learning and develop a principled formulation for trajectory-level valuation. 

\section{Preliminaries}\label{sec:pre}
Below, we elaborate on the preliminaries in terms of reinforcement learning and influence functions.

\subsection{Reinforcement Learning}
A standard reinforcement learning framework is modeled as a Markov Decision Process, defined by a tuple $(\mathcal{S}, \mathcal{A}, P, r, \gamma)$, where $\mathcal{S}$ and $\mathcal{A}$ denote the state and action spaces, $P(s' \mid s,a)$ is the transition dynamics, $r(s,a)$ is the reward function, and $\gamma \in [0,1)$ is the discount factor. A policy $\pi_\theta(a \mid s)$, parameterized by $\theta$, maps states to action distributions. The goal of reinforcement learning is to maximize the expected discounted return:
\begin{equation}
J(\theta) = \mathbb{E}_{\tau \sim \pi_\theta} \left[\sum_{t=0}^{T} \gamma^t r(s_t, a_t)\right],
\end{equation}
where $\tau$$=$$(s_0, a_0, \dots, s_T)$ denotes a trajectory.

A common approach to optimizing this objective is via policy gradient methods, which estimate the gradient of $J(\theta)$ as:
\begin{equation}
\nabla_\theta J(\theta) = \mathbb{E}_{\tau \sim \pi_\theta} \left[\sum_{t=0}^{T} \nabla_\theta \log \pi_\theta(a_t \mid s_t) \cdot A^{\pi}(s_t, a_t)\right],
\end{equation}
where $A^{\pi}(s,a)$ is the advantage function, measuring the relative contribution of an action compared to a baseline policy.

In practice, optimizing this objective in high-dimensional settings requires stable and scalable algorithms. Importantly, many modern approaches operate at the level of complete trajectories, where each rollout---i.e., a trajectory generated by the policy during interaction with the environment---serves as a fundamental training unit. We next introduce several representative methods in this regime, which form the basis of our study.

% \begin{equation}
% r_t(\theta) = \frac{\pi_\theta(a_t \mid s_t)}{\pi_{\theta_{\text{old}}}(a_t \mid s_t)}
% \end{equation}
\textbf{Proximal Policy Optimization (PPO)}.
Proximal Policy Optimization (PPO)~\citep{schulman2017proximal} is a widely used policy gradient method that stabilizes training via a clipped surrogate objective. Let $r_t(\theta)$$=$$\pi_\theta(a_t \mid s_t)/\pi_{\theta_{\text{old}}}(a_t \mid s_t)$ denote the importance sampling ratio. The PPO objective is:
\begin{equation}
\mathcal{L}_{\text{PPO}}(\theta) = \mathbb{E}_t \left[
\min \left(
r_t(\theta) A_t,\ 
\text{clip}(r_t(\theta), 1 - \epsilon, 1 + \epsilon) A_t
\right)
\right].
\end{equation}
This clipping mechanism prevents excessively large policy updates and improves training stability.

% \begin{equation}
% \tilde{A}_i = \frac{r_i - \mu_{\mathcal{G}}}{\sigma_{\mathcal{G}}},
% \end{equation}

\textbf{Group Relative Policy Optimization (GRPO)}.
Group Relative Policy Optimization (GRPO)~\citep{shao2024deepseekmath} extends PPO by replacing absolute advantage estimates with relative advantages computed within a group of sampled trajectories. Given a group $\mathcal{G}$$=$$\{\tau_i\}_{i=1}^K$, GRPO normalizes rewards within the group by $\tilde{A}_i$$=$$(r_i - \mu_{\mathcal{G}})/\sigma_{\mathcal{G}},$ where $\mu_{\mathcal{G}}$ and $\sigma_{\mathcal{G}}$ are the mean and standard deviation of rewards in the group. The objective of GRPO becomes:
\begin{equation}
\mathcal{L}_{\text{GRPO}}(\theta) = \mathbb{E}_{i,t} \left[
\min \left(
r_{i,t}(\theta) \tilde{A}_i,\ 
\text{clip}(r_{i,t}(\theta), 1 - \epsilon, 1 + \epsilon) \tilde{A}_i
\right)
\right].
\end{equation}
By leveraging relative comparisons, GRPO reduces reliance on value function estimation and improves robustness in scenarios such as RLHF.

\textbf{Direct Preference Optimization (DPO)}.
Direct Preference Optimization (DPO)~\citep{rafailov2023direct} formulates policy learning directly from preference data without explicit reward modeling. Given a pair of responses $(y^+, y^-)$ for a prompt $x$, where $y^+$ is preferred over $y^-$, DPO optimizes:
\begin{equation}
\mathcal{L}_{\text{DPO}}(\theta) = - \mathbb{E}_{(x, y^+, y^-)} \left[
\log \sigma \left(
\beta \left(
\log \frac{\pi_\theta(y^+ \mid x)}{\pi_{\text{ref}}(y^+ \mid x)} -
\log \frac{\pi_\theta(y^- \mid x)}{\pi_{\text{ref}}(y^- \mid x)}
\right)
\right)
\right],
\end{equation}
where $\pi_{\text{ref}}$ is a reference policy and $\beta$ controls the sharpness. DPO can be interpreted as implicitly optimizing a KL-regularized reward objective while bypassing explicit reward model training.

In sum, PPO provides a stable on-policy optimization framework, GRPO enhances it with relative trajectory comparisons, and DPO extends RL to preference-based optimization without explicit rewards. These methods form the backbone of modern reinforcement learning, especially in large-scale language model post-training.

\subsection{Influence Function}
The effect of an individual training sample can be characterized by infinitesimally perturbing its contribution to the training objective and tracing the resulting change in model behavior. Given a model parameterized by $\theta$, let $z$ denote an individual training sample and the empirical risk minimization objective be defined as $\hat{\theta}$$=$$\arg\min_{\theta\in\Theta}\frac{1}{n}\sum_{i=1}^{n}\ell(z_i;\theta)$. Following the classical influence-function formulation of \citet{koh2017understanding}, the effect of removing a training sample $z_j$ on a validation objective can be approximated by infinitesimally perturbing its contribution to the training objective. This yields the following influence score:

\begin{equation}
\label{eq:influence}
\mathcal{I}(z_j;\hat{\theta})
=
\nabla f(\mathcal{V};\hat{\theta})^{\top}
\mathbf{H}_{\hat{\theta}}^{-1}
\nabla\ell(z_j;\hat{\theta}),
\end{equation}

where $\mathcal{V}$ denotes the validation set, $f(\cdot)$ represents the impact function that specifies the evaluation objective on $\mathcal{V}$, $\nabla\ell(z_j;\hat{\theta})$ denotes the gradient contribution of sample $z_j$, and $\mathbf{H}_{\hat{\theta}}$$=$$\sum_{i=1}^{n}\nabla^2\ell(z_i;\hat{\theta})$ is the Hessian matrix. This formulation provides a principled measure of how individual training samples affect model behavior through their contribution to the optimization process.

% \newpage
\section{Method}
\label{sec:method}
In this section, we first highlight three critical challenges when adapting traditional influence functions to reinforcement learning. We then present Dynamic Trajectory Valuation (DTV), a unified framework that overcomes these issues and seamlessly integrates across various reinforcement learning paradigms. Lastly, we offer an in-depth analysis of DTV by exploring its trajectory decomposition and leave-one-out extension.

\subsection{Challenges of Influence Functions in Reinforcement Learning}
Influence functions, as a crucial tool for data-centric learning, have been extensively studied in supervised learning~\citep{koh2017understanding}; however, their application to reinforcement learning remains largely underexplored.  Extending these principles to reinforcement learning is non-trivial due to several fundamental differences:
\vspace{-2mm}

\begin{itemize}[wide=10pt, leftmargin=*]
\item \textit{Input Trajectories}:
Supervised learning assumes fixed training and validation datasets, enabling data valuation through each sample's influence on validation performance. In contrast, reinforcement learning generates trajectories online through environment interactions, with no fixed training set for valuation, rendering conventional valuation methods inapplicable.

\item \textit{Impact Function}:
Classical influence functions rely on an explicit evaluation function defined on a validation set to quantify the contribution of individual samples. In reinforcement learning, trajectory usefulness is inherently policy-dependent and shaped by temporal credit assignment and reward dynamics, making such definitions elusive.

\item \textit{Dynamic Mechanism}:
Conventional data valuation typically relies on static, post-hoc analysis. Reinforcement learning, however, requires dynamic, online valuation as trajectories evolve, raising the challenge of estimating trajectory impact efficiently without incurring prohibitive computational overhead.

\end{itemize}
\vspace{-1mm}

Recent efforts~\citep{hu2025snapshot,shu2026learning,li2026learnalign,yang2026gradalign} extend data valuation and selection to reinforcement learning. Existing methods rely on setting-specific designs, including replay buffers and surrogate impact functions for PPO, verifiable-feedback-based learnability, and trusted validation gradients. However, they lack a unified and principled formulation across reinforcement learning paradigms, raising concerns about their generalizability.

\subsection{Dynamic Trajectory Valuation}
\label{subsec:dtv}

In this paper, we consider a principled formulation for general reinforcement learning by addressing the above three challenges in a unified framework. If we take a close look at Eq.~\eqref{eq:influence}, the first and second terms are constants across candidate samples, even given the unknown validation set and impact function. Therefore, the sample gradient is decisive for trajectory valuation, which casts the trajectory valuation problem into a sample gradient analysis problem. Suppose that learning is effective, beneficial samples should constitute the majority within a mini-batch, which leads to the following assumption. 

\begin{assumption}[Majority Gradient Alignment]
\label{assump:majority_alignment}
Within a mini-batch, the majority of sample gradients collectively provide a locally informative reference direction for the current update. Accordingly, a sample whose gradient is negatively aligned with this reference direction is regarded as detrimental to the current optimization step.
\end{assumption}

Under Assumption~\ref{assump:majority_alignment}, we propose Dynamic Trajectory Valuation (DTV) to dynamically identify detrimental training units during optimization. A \textit{training unit} denotes the basic entity for which a per-unit gradient contribution is computed, with its form depending on the underlying learning paradigm: a trajectory segment in PPO, a completion trajectory generated for a prompt in GRPO, and a prompt with a chosen--rejected response pair in DPO. Given a mini-batch $\mathcal{B}$$=$$\{z_j\}_{j=1}^b$, let $g_j^{(t)}$$=$$\nabla_\theta \ell(z_j;\theta^{(t)})$ denote the gradient contribution of training unit $z_j$, where $\ell(z_j;\theta)$ is the training loss under the underlying algorithm. We define the DTV score as
\begin{equation}
\label{eq:dtv}
\mathcal{I}^{\mathrm{DTV}}(z_j;\theta^{(t)})
=
\frac{1}{b}\sum_{i=1}^{b}
\bigl(g_i^{(t)}\bigr)^{\top} g_j^{(t)}.
\end{equation}

DTV requires no separate impact function or predefined validation set. Instead, its reference direction is derived directly from the current mini-batch gradients. Training units with negative DTV scores are identified as alignment-opposing and filtered out dynamically before executing the optimizer update.

\textbf{Intuition and Dynamics.}
Measuring alignment in the gradient space directly reflects how each training unit impacts parameter trajectory, avoiding heuristic assumptions in the original input/feature space. While prior work~\citep{chhabra2024outlier} considers gradient-based outlier analysis for static data valuation in classification (e.g., using Isolation Forests), DTV addresses the dynamic nature of reinforcement learning. By leveraging the majority gradient as an online reference, DTV naturally adapts to evolving policy checkpoints without requiring global convergence or local convexity assumptions~\citep{wang2024data, yang2025layer}.

\textbf{Computational Efficiency.} A key advantage of our DTV is that it enables direct and efficient valuation without additional gradient approximations. Prior implementations rely on batch-level gradients and require surrogate techniques such as ghost influence or layer-wise estimation~\citep{wang2024data, yang2025layer}. In contrast, our JAX~\citep{jax2018github} implementation leverages vectorized automatic differentiation to directly compute training-unit-level gradients, enabling efficient evaluation of Eq.~\eqref{eq:dtv} and seamless integration into diverse reinforcement learning pipelines.

\subsection{DTV Decomposition and Leave-One-Out Extension}
\label{subsec:dtv_loo}

To clarify the information captured by DTV, we decompose the score in Eq.~\eqref{eq:dtv} into the contribution from the evaluated training unit itself and its interactions with the remaining mini-batch. Expanding the mini-batch mean gives $\mathcal{I}^{\mathrm{DTV}}(z_j;\theta^{(t)})$$=$$\frac{1}{b}\|g_j^{(t)}\|^2$$+$$\frac{1}{b}\sum_{i\neq j}(g_j^{(t)})^\top g_i^{(t)}$$=$$s_j^{(t)}$$+$$c_j^{(t)}$, where $s_j^{(t)}$$=$$\frac{1}{b}\|g_j^{(t)}\|^2$ and $c_j^{(t)}$$=$$\frac{1}{b}\sum_{i\neq j}(g_j^{(t)})^\top g_i^{(t)}$ denote the \textit{self-term} and \textit{cross-term}, respectively. The self-term captures self-alignment, whereas the cross-term measures aggregate gradient agreement with the remaining training units. 

\begin{remark}[Self-Protection Effect]
\label{rem:self_protection}
Since $s_j^{(t)}$$\geq$$0$, DTV filters a training unit only when $c_j^{(t)}$$<$$-s_j^{(t)}$. A sufficiently large self-term can therefore offset a negative cross-term and retain a training unit even when it is strongly negatively aligned with the remaining mini-batch, an effect we refer to as self-protection. Nevertheless, the self-term should not be interpreted as universally detrimental, as its effect may depend on the underlying optimization setting.
\end{remark}

The self-protection effect motivates a leave-one-out extension that isolates the cross-term by excluding the gradient itself from the reference direction. Specifically, we average over the remaining $b$$-$$1$ units and define DTV-Loo as
\begin{equation}
\label{eq:dtv_loo}
\mathcal{I}^{\mathrm{DTV\text{-}Loo}}(z_j;\theta^{(t)})=\frac{1}{b-1}\sum_{i\neq j}\bigl(g_i^{(t)}\bigr)^\top g_j^{(t)}.
\end{equation}
Using the decomposition above, DTV-Loo can equivalently be written as $\mathcal{I}^{\mathrm{DTV\text{-}Loo}}$$=$$\frac{b}{b-1}\bigl(\mathcal{I}^{\mathrm{DTV}}$$-$$s_j^{(t)}\bigr)$. Consequently, DTV-Loo removes the self-term while maintaining exact leave-one-out normalization. Its valuation thus relies solely on the cross-term, effectively preventing a single harmful sample from dominating the mini-batch evaluation.

More generally, DTV and DTV-Loo can be unified into a broader $\mathrm{DTV\text{-}}\lambda$ family ($\lambda\in [0,1]$) by weighting the self-term. Here, $\lambda$$=$$1$ recovers DTV, whereas $\lambda$$=$$0$ completely removes the self-term to yield DTV-Loo. Despite this broader parameterization, we retain DTV and DTV-Loo as the practical instantiations, as both already achieve strong empirical performance without requiring additional tuning of $\lambda$. Empirically, DTV-Loo performs better than DTV in most evaluated settings, whereas DTV can be more effective under specific regimes, demonstrating their complementary strengths across different optimization scenarios and providing practical guidelines for method selection. Algorithm~\ref{alg:dtv_filtering} summarizes the filtering procedure.

% More generally, DTV and DTV-Loo represent two endpoints that differ in the weight assigned to the evaluated gradient itself when constructing the reference direction. We characterize this relation through a unified DTV-$\lambda$ family:

% \begin{equation}
% \label{eq:dtv_lambda}
% \mathcal{I}^{\mathrm{DTV}_{\lambda}}(z_j;\theta^{(t)})= \frac{\lambda\|g_j^{(t)}\|^2+\sum_{i\neq j}\bigl(g_j^{(t)}\bigr)^\top g_i^{(t)}}{b-1+\lambda}, \qquad \lambda\in[0,1].\td
% \end{equation}

% Here, $\lambda$ controls the weight of the evaluated gradient itself in constructing the reference direction. Under the decomposition above, $\mathcal{I}^{\mathrm{DTV}_{\lambda}}$ can be written as $\frac{b}{b-1+\lambda}\bigl(\lambda s_j^{(t)}+c_j^{(t)}\bigr)$, making its effect on the self-term explicit. Accordingly, $\lambda=1$ exactly recovers DTV, whereas $\lambda=0$ recovers DTV-Loo. We use this family only as an analytical unification and evaluate the two parameter-free endpoints, DTV and DTV-Loo, in our experiments.

 % The valuation method $M$ determines whether the evaluated gradient contributes to the construction of the reference direction used to compute each training-unit score.

\begin{algorithm}[H]
\small
\caption{Dynamic Training-Unit Filtering with DTV and DTV-Loo}
\label{alg:dtv_filtering}
\begin{algorithmic}[1]

\Require Parameters $\theta^{(0)}$; steps $T$; valuation method $M\in\{\mathrm{DTV},\mathrm{DTV\text{-}Loo}\}$; per-unit objective $\ell$
\Ensure Updated model parameters $\theta^{(T)}$

\For{$t=0,\ldots,T-1$}
\State Obtain training units $\mathcal{B}^{(t)}$$=$$\{z_j\}_{j=1}^{b}$
\State Compute $g_j^{(t)}\gets\nabla_\theta\ell(z_j;\theta^{(t)})$ for all $z_j\in\mathcal{B}^{(t)}$
\State Compute valuation scores $\mathcal{I}_j^{(t)}\gets\mathcal{I}^{M}(z_j;\theta^{(t)})$ for all $j$
\State Retain $\mathcal{B}_{+}^{(t)}\gets\{z_j\in\mathcal{B}^{(t)}:\mathcal{I}_j^{(t)}\geq0\}$
\State Update $\theta^{(t+1)}\gets\operatorname{Update}\bigl(\theta^{(t)},\mathcal{B}_{+}^{(t)}\bigr)$
\EndFor

\State \Return $\theta^{(T)}$

\end{algorithmic}
\end{algorithm}

\FloatBarrier

% \newpage
\section{Experiments}
\label{sec:experiments}
To evaluate the versatility and practical applicability of DTV and DTV-Loo, we apply the proposed methods across diverse reinforcement learning and LLM post-training settings. Specifically, we study online trajectory optimization with PPO, group-relative optimization with GRPO, and pairwise preference optimization with DPO, spanning different training-unit granularities, including trajectory segments, completion groups, and preference pairs. Beyond evaluating final performance, we further investigate the self-protection effect underlying the distinction between DTV and DTV-Loo, optimization dynamics, and data efficiency. All experiments follow a JAX-based~\citep{jax2018github} implementation, with the GRPO and DPO pipelines further built on Google's open-source Tunix library~\citep{tunix2025}, which we extend with vectorized training-unit-level gradient computation and dynamic filtering.

\subsection{DTV for Online Trajectory Valuation with PPO}
\label{subsec:ppo}

% \vspace{8mm}
\begin{wrapfigure}[21]{r}{0.22\textwidth}
    \centering
    \vspace{-12mm}
    \begin{subfigure}[t]{\linewidth}
        \centering
        \includegraphics[width=\linewidth]{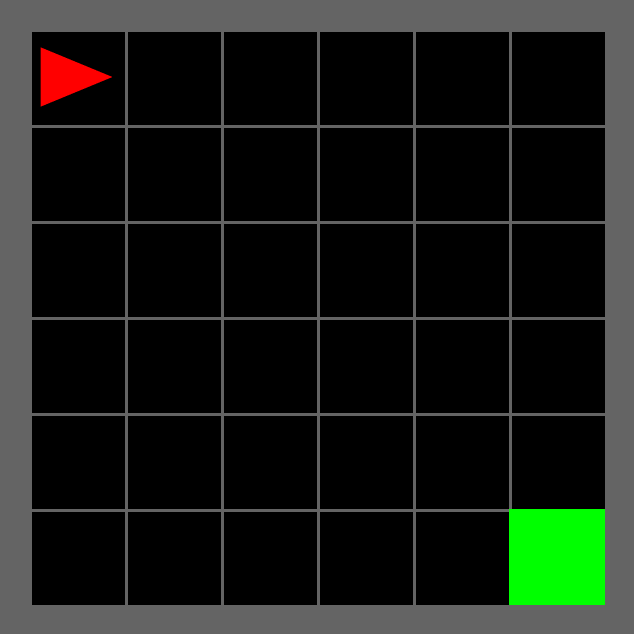}
        \vspace{-4mm}
        \caption{\textit{Empty-8x8}}
        \label{fig:ppo_empty_env}
    \end{subfigure}

    \vspace{2mm}

    \begin{subfigure}[t]{\linewidth}
        \centering
        \includegraphics[width=\linewidth]{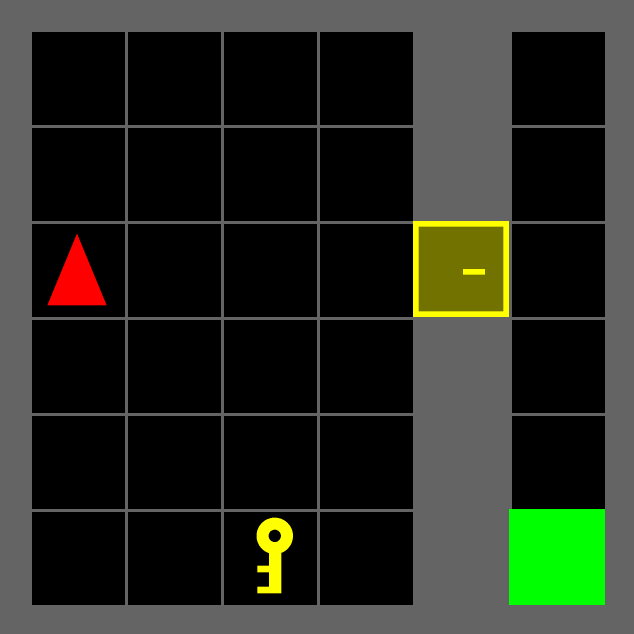}
        \vspace{-4mm}        
        \caption{\textit{DoorKey-8x8}}
        \label{fig:ppo_doorkey_env}
    \end{subfigure}
    \vspace{-2mm}
    \caption{\textit{MiniGrid} tasks.}
    \label{fig:ppo_environments}
    \vspace{-2mm}
\end{wrapfigure}

\textbf{Experimental Setup.} We compare DTV and DTV-Loo with Vanilla PPO and Iterative Influence-Based Filtering (IIF)~\citep{hu2025snapshot} in the PPO setting. Vanilla PPO performs standard PPO updates without filtering. IIF serves as an influence-based filtering baseline by estimating the impact of individual transitions. In contrast, DTV and DTV-Loo perform adaptive trajectory-level filtering, where rollouts are partitioned into trajectory-level training units, with boundaries determined by episode termination or the rollout boundary. Following Algorithm~\ref{alg:dtv_filtering}, trajectory units with non-negative valuation scores are retained, while filtering a unit excludes all of its constituent transitions from the subsequent PPO update.

\begin{figure}[t]
    \centering
    \begin{subfigure}[t]{0.33\linewidth}
        \centering       \includegraphics[width=\linewidth] {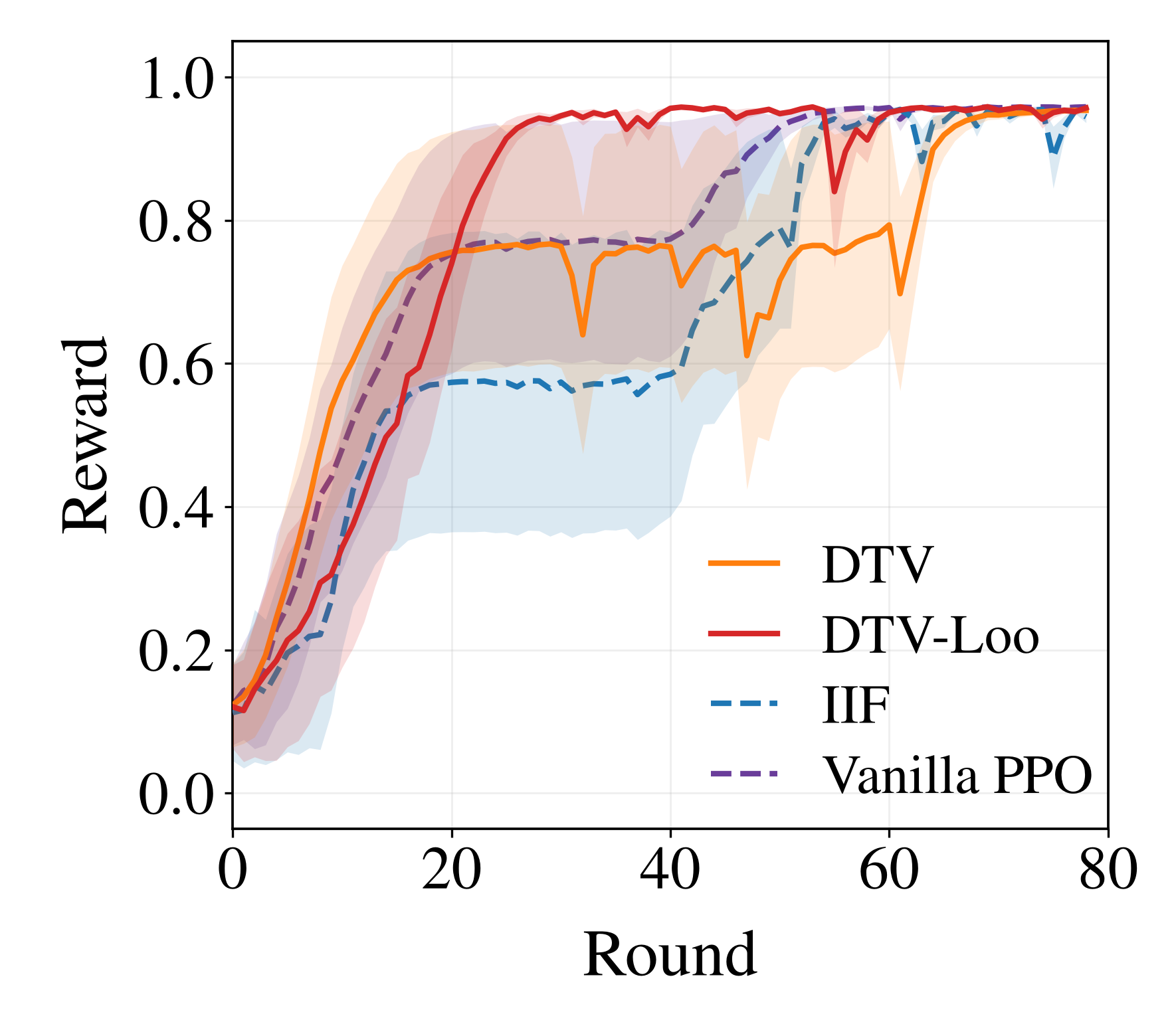}
        \vspace{-7mm}
        \captionsetup{margin={5mm,0mm}}
    \caption{\textit{Empty-8x8}}\label{fig:ppo_empty_result}
    \end{subfigure}\hfill%
    \begin{subfigure}[t]{0.33\linewidth}
        \centering
    \includegraphics[width=\linewidth]{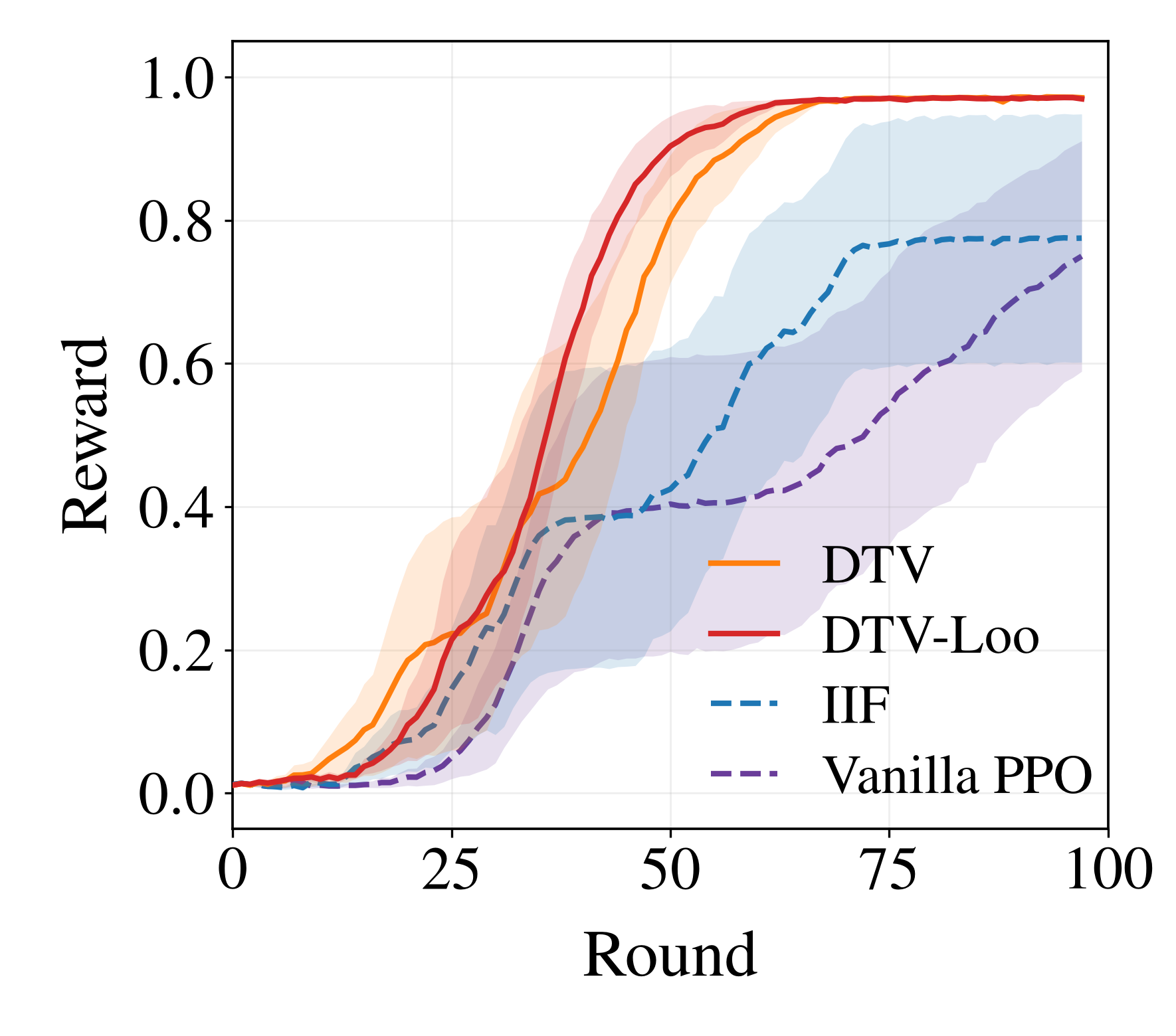}
        \vspace{-7mm}
        \captionsetup{margin={4mm,0mm}}\caption{\textit{DoorKey-8x8}}
\label{fig:ppo_doorkey_result}
    \end{subfigure}\hfill%
    \begin{subfigure}[t]{0.33\linewidth}
        \centering
\includegraphics[width=\linewidth]
{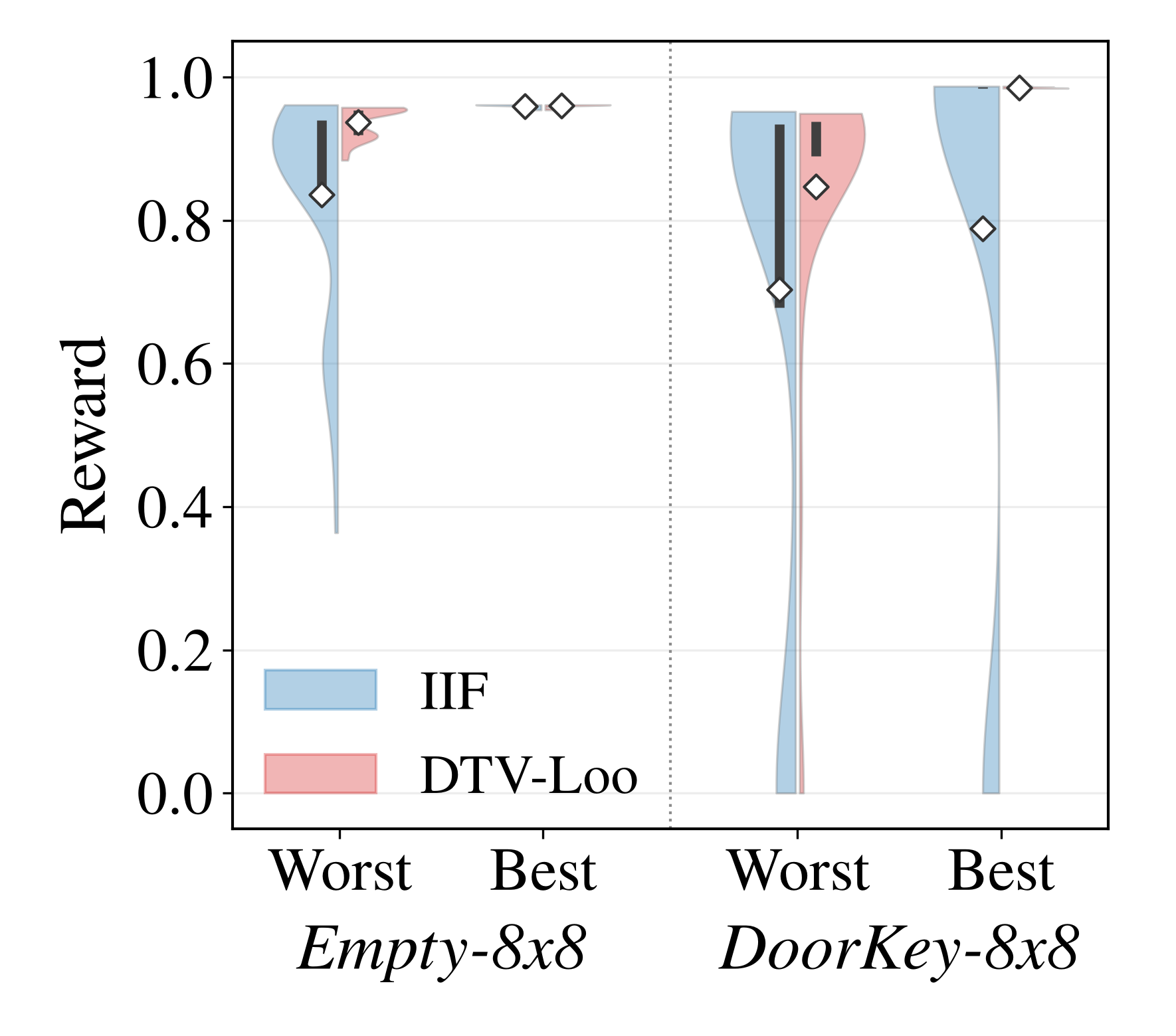}       \vspace{-7mm}
        \captionsetup{margin={6mm,0mm}}
        \caption{Worst/Best returns}
\label{fig:ppo_best_worst}
    \end{subfigure}
    \vspace{-1mm}
    \caption{
    \textbf{PPO results on \textit{MiniGrid}.}
    (a--b) Mean test reward over PPO rounds across five seeds; shaded regions show $\pm1$ SEM.
    (c) At the final checkpoint, the lowest and highest 20 returns from $1{,}000$ evaluation episodes per seed are pooled across five seeds as the \emph{Worst} and \emph{Best} subsets.
    }\vspace{-4mm}
    \label{fig:ppo_results}
\end{figure}

We conduct PPO experiments on two partially observable \textit{MiniGrid} environments~\citep{MinigridMiniworld23}. \textit{Empty-8x8} follows the setting used by IIF~\citep{hu2025snapshot}, enabling a direct comparison on sparse-reward navigation. \textit{DoorKey-8x8}, in contrast, requires the agent to collect a key, unlock a door, and reach the goal, introducing longer-horizon dependencies and harder exploration. 
% All methods use identical architectures, interaction budgets, and PPO hyperparameters within each environment. \textit{Empty-8x8} and \textit{DoorKey-8x8} are trained for $160$K and $1$M environment steps, respectively. For all filtering methods, filtering starts immediately on \textit{Empty-8x8} and after five warmup rounds on \textit{DoorKey-8x8}. 
We report the mean over five seeds, with shaded regions indicating $\pm1$ standard error of the mean (SEM); each checkpoint is evaluated over $1,000$ episodes. Full architectures, hyperparameters, and filtering configurations are provided in Appendix~\ref{app:ppo}.

\textbf{Main Results.} Figure~\ref{fig:ppo_results}(a)--(b) show that DTV-based trajectory filtering accelerates PPO convergence across both \textit{MiniGrid} tasks. On \textit{Empty-8x8}, DTV-Loo converges substantially earlier than the other methods, although all methods ultimately attain comparable returns. On the more challenging \textit{DoorKey-8x8}, DTV-Loo exhibits slightly faster early-stage convergence than DTV; thereafter, both variants approach the reward ceiling and maintain a clear advantage over IIF and Vanilla PPO under the same interaction budget. Notably, the performance margin over both baselines is substantially larger on \textit{DoorKey-8x8}, underscoring the robustness of the gains achieved by DTV-based filtering in the more challenging environment. 

At the final checkpoint, Figure~\ref{fig:ppo_results}(c) further shows that DTV-Loo yields higher returns than IIF in both the Worst and Best subsets across the two environments. These improvements are consistent with the gradient-alignment perspective in Section~\ref{sec:method}: by removing trajectory units that conflict with the dominant update direction, DTV-based filtering reduces detrimental optimization interference while preserving useful training signals. As a result, the policy achieves more stable improvement despite using fewer effective training units. Together, these results indicate that DTV-based trajectory filtering accelerates PPO convergence and improves the consistency of final-policy performance.

Overall, the PPO results establish two consistent benefits of DTV-based filtering: faster convergence and improved final-policy performance. The convergence advantage is further supported in our later DPO experiments, where DTV-Loo reaches the same performance target substantially faster. The performance benefit is also observed across the subsequent GRPO and DPO settings. Given the extensive prior study of PPO, we focus on two relatively standard benchmark settings here. We reserve a deeper analysis of the mechanisms underlying DTV and DTV-Loo for the more complex GRPO and DPO experiments that follow. 

\begin{table*}[t]
\centering
% \footnotesize
\small
\setlength{\tabcolsep}{8.0 pt}
\caption{
\textbf{GRPO results on \textit{GSM8K} under clean and mismatch settings.}
$\Delta$ Acc. is measured relative to the shared pre-trained model. Results are reported as mean $\pm$ std.\ over five seeds, with the best result shown in bold. Given the consistent margins, we omit additional pairwise $t$-tests.
}
\vspace{-2mm}
\renewcommand{\arraystretch}{1.08}
\begin{tabular}{@{}lcccccc@{}}
% \begin{tabular}{lcccccc}
\toprule
& \multicolumn{3}{c}{\textit{Clean}}
& \multicolumn{3}{c}{\textit{Mismatch-20\%}} \\
\cmidrule(lr){2-4}
\cmidrule(lr){5-7}
Method
& Acc. $\uparrow$
& $\Delta$ Acc. $\uparrow$
& Partial $\uparrow$
& Acc. $\uparrow$
& $\Delta$ Acc. $\uparrow$
& Partial $\uparrow$ \\
\midrule

Pre-trained
& 0.4723 $\pm$ 0.0000
& 0.0000
& 0.4996 $\pm$ 0.0000
& 0.4723 $\pm$ 0.0000
& 0.0000
& 0.4996 $\pm$ 0.0000 \\
\cmidrule(lr){1-7}

Vanilla GRPO
& 0.4828 $\pm$ 0.0077
& 0.0105
& 0.5114 $\pm$ 0.0089
& 0.4731 $\pm$ 0.0094
& 0.0008
& 0.5001 $\pm$ 0.0089 \\

Random 5\%
& 0.5207 $\pm$ 0.0341
& 0.0484
& 0.5442 $\pm$ 0.0307
& 0.4103 $\pm$ 0.2026
& -0.0620
& 0.4255 $\pm$ 0.2086 \\

Random 10\%
& 0.4511 $\pm$ 0.1175
& -0.0212
& 0.4760 $\pm$ 0.1127
& 0.5133 $\pm$ 0.0431
& 0.0409
& 0.5318 $\pm$ 0.0455 \\

Reward 5\%
& 0.4516 $\pm$ 0.2104
& -0.0208
& 0.4729 $\pm$ 0.2181
& 0.4443 $\pm$ 0.1691
& -0.0281
& 0.4628 $\pm$ 0.1733 \\

Reward 10\%
& 0.4973 $\pm$ 0.1128
& 0.0250
& 0.5193 $\pm$ 0.1094
& 0.3777 $\pm$ 0.1981
& -0.0946
& 0.4012 $\pm$ 0.2000 \\

LearnAlign~(\citeyear{li2026learnalign})
& 0.5014 $\pm$ 0.0170
& 0.0291
& 0.5281 $\pm$ 0.0160
& 0.4863 $\pm$ 0.0069
& 0.0140
& 0.5130$\pm$0.0083 \\

GradAlign~(\citeyear{yang2026gradalign})
& 0.4876 $\pm$ 0.0040
& 0.0153
& 0.5154 $\pm$ 0.0046
& 0.4775 $\pm$ 0.0064
& 0.0052
& 0.5060 $\pm$ 0.0051 \\

\textbf{DTV (Ours)}
& \textbf{0.5483 $\pm$ 0.0075}
& \textbf{0.0760}
& \textbf{0.5742 $\pm$ 0.0099}
& \textbf{0.5413 $\pm$ 0.0139}
& \textbf{0.0690}
& \textbf{0.5660 $\pm$ 0.0123} \\

DTV-Loo (Ours)
& 0.5416 $\pm$ 0.0141
& 0.0693
& 0.5647 $\pm$ 0.0146
& 0.5234 $\pm$ 0.0103
& 0.0511
& 0.5460 $\pm$ 0.0076 \\

\bottomrule
\end{tabular}
\label{tab:grpo_gsm8k_results}\vspace{-4mm}
\end{table*}

\subsection{DTV for Group-Relative Completion Valuation with GRPO}

Next, we examine DTV and DTV-Loo in the GRPO setting on two mathematical reasoning settings of different difficulty, ranging from grade-school reasoning on \textit{GSM8K}~\citep{cobbe2021gsm8k} to more challenging competition-level reasoning on the American Invitational Mathematics Examination (AIME). GRPO generates multiple completion trajectories for each prompt and optimizes them using group-relative advantages, providing a natural setting for completion-level valuation and filtering. Across both benchmarks, DTV and DTV-Loo use policy-loss gradients to value individual completion trajectories relative to other completions generated for the same prompt. We further analyze the role of the self-term on \textit{GSM8K} to better understand the performance differences between DTV and DTV-Loo under group-relative optimization.

\textbf{Experimental Setup.} 
We evaluate DTV and DTV-Loo against Vanilla GRPO and two fixed-ratio filtering baselines. Vanilla GRPO performs the standard group-relative update without filtering. All filtering methods operate on individual completions within each prompt group. Random Filtering selects completions independently of their reward signals, whereas Reward-based Filtering removes those with the lowest observed group-relative advantages. DTV and DTV-Loo instead determine which completions to retain adaptively based on within-group policy-gradient alignment, without prescribing a filtering rate. On \textit{GSM8K}, we conduct the full comparison with Random and Reward-based Filtering at expected rates of 5\% and 10\%. On the substantially more computationally demanding \textit{AIME} setting, we focus on the three core methods, Vanilla GRPO, DTV, and DTV-Loo. 

For \textit{GSM8K}~\citep{cobbe2021gsm8k}, we fine-tune Gemma-3-1B-IT~\citep{gemmateam2025gemma3technicalreport} with LoRA under \textit{Clean} and \textit{Mismatch-20\%}, where the latter perturbs the within-group reward ranking while keeping evaluation data clean. We restrict the main corruption study to \textit{Mismatch-20\%}, since extending the same rank-reversal perturbation to 40\% of prompt groups would substantially distort the within-group relative signal that directly drives the GRPO update. All methods use the same training budget and are evaluated over five matched seeds on the full \textit{GSM8K} test set. We report exact-answer accuracy (Acc.), the improvement over the shared pre-trained model ($\Delta$ Acc.), and partial accuracy (Partial), which counts numerical predictions within 10\% relative error of the ground-truth answer.

For the more challenging \textit{AIME 2024}~\citep{huggingfaceh4_2024_aime}, we full-parameter fine-tune DeepSeek-R1-Distill-Qwen-1.5B~\citep{deepscaler2025} on the DeepScaleR-Preview-Dataset~\citep{deepscaler2025}. Each optimization step samples eight completions per prompt, with binary mathematical correctness used as the reward for group-relative advantage estimation. Training uses a maximum response length of 8,192 tokens, whereas final evaluation uses 32,768 tokens to reduce truncation, with 16 responses sampled for each of the \textit{AIME 2024} problems. We report Avg@16, the average correctness over all sampled responses; Pass@16, the fraction of problems solved by at least one response; Maj@16, the majority-vote accuracy over the 16 responses; Extractable, the fraction of responses with a successfully parsed final answer; and Acc. given extractable, the accuracy conditioned on successful answer extraction. Full training, generation, filtering, and evaluation details for both benchmarks are provided in Appendix~\ref{app:grpo}.

% \vspace{-7mm}
\begin{figure}[!tb]
\centering
\begin{subfigure}[t]{0.33\textwidth}
    \centering
    \includegraphics[width=\linewidth]{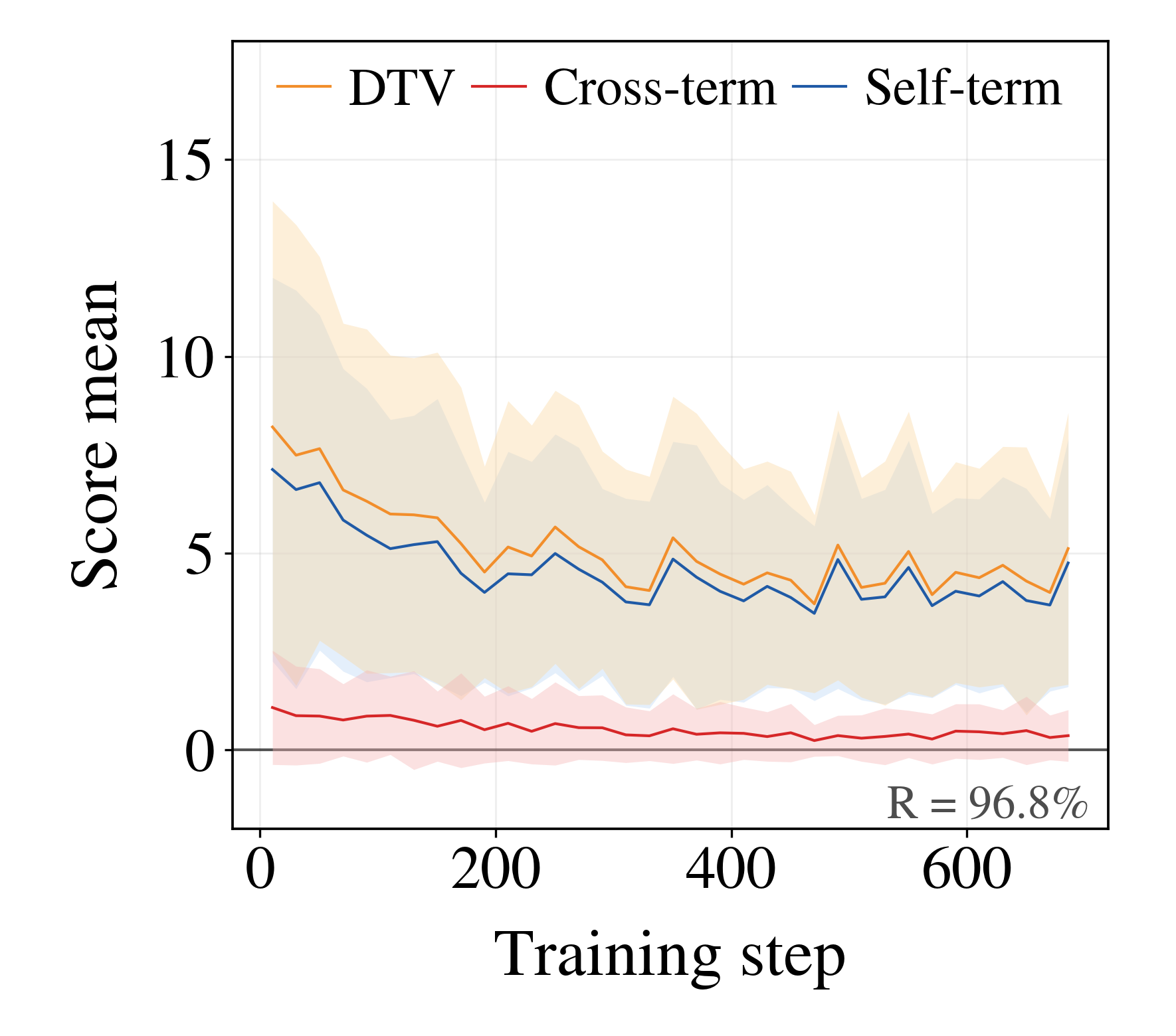}
    \vspace{-7mm}
    \captionsetup{margin={5mm,0mm}}
    \caption{Score decomposition}
    \label{fig:gsm8k_score_decomposition}
\end{subfigure}
\hfill
\begin{subfigure}[t]{0.33\textwidth}
    \centering
    \includegraphics[width=\linewidth]{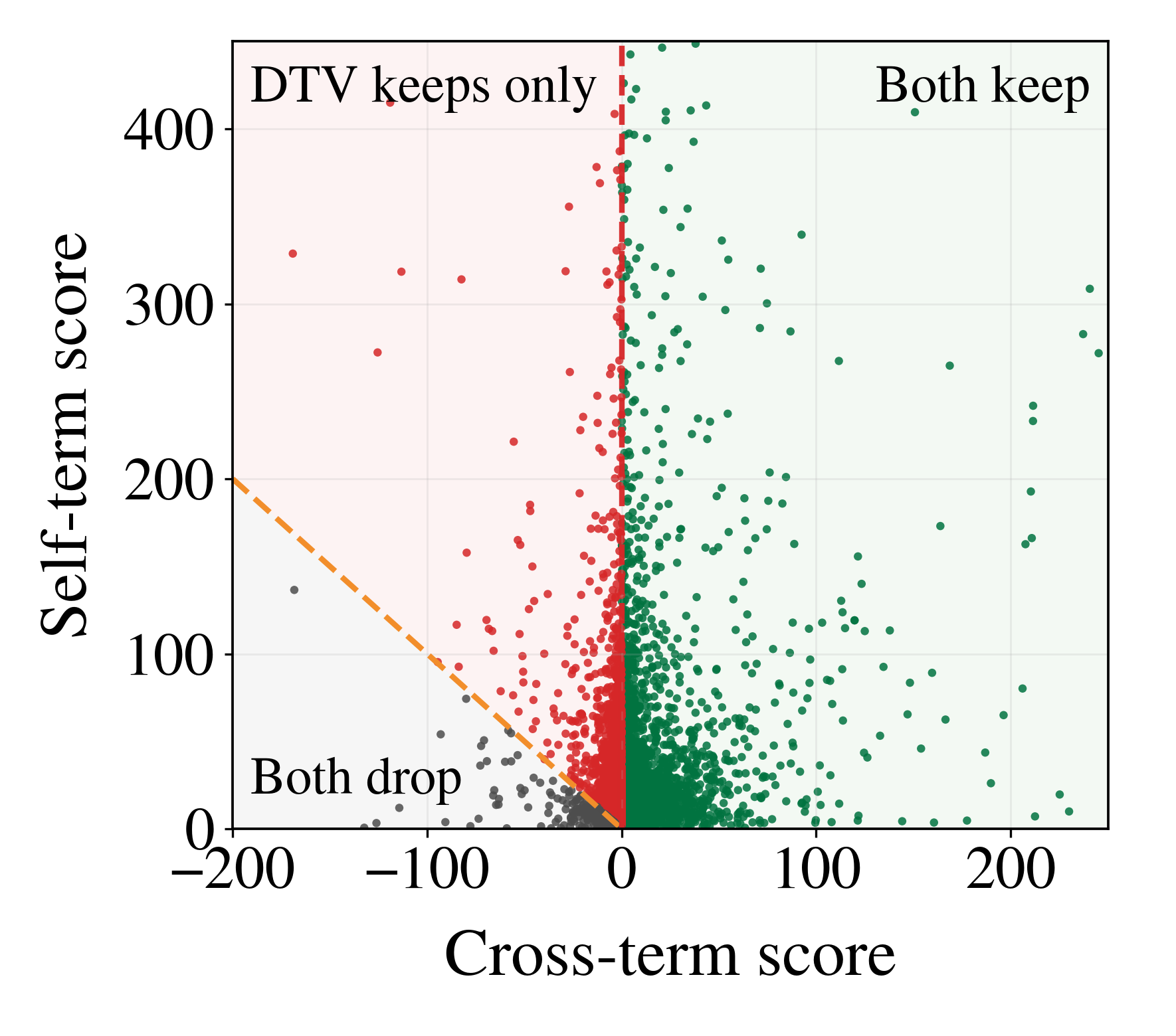}
    \vspace{-7mm}
    \captionsetup{margin={5mm,0mm}}
    \caption{Filtering regions}
    \label{fig:gsm8k_filtering_regions}
\end{subfigure}
\hfill
\begin{subfigure}[t]{0.33\textwidth}
    \centering
    \includegraphics[width=\linewidth]{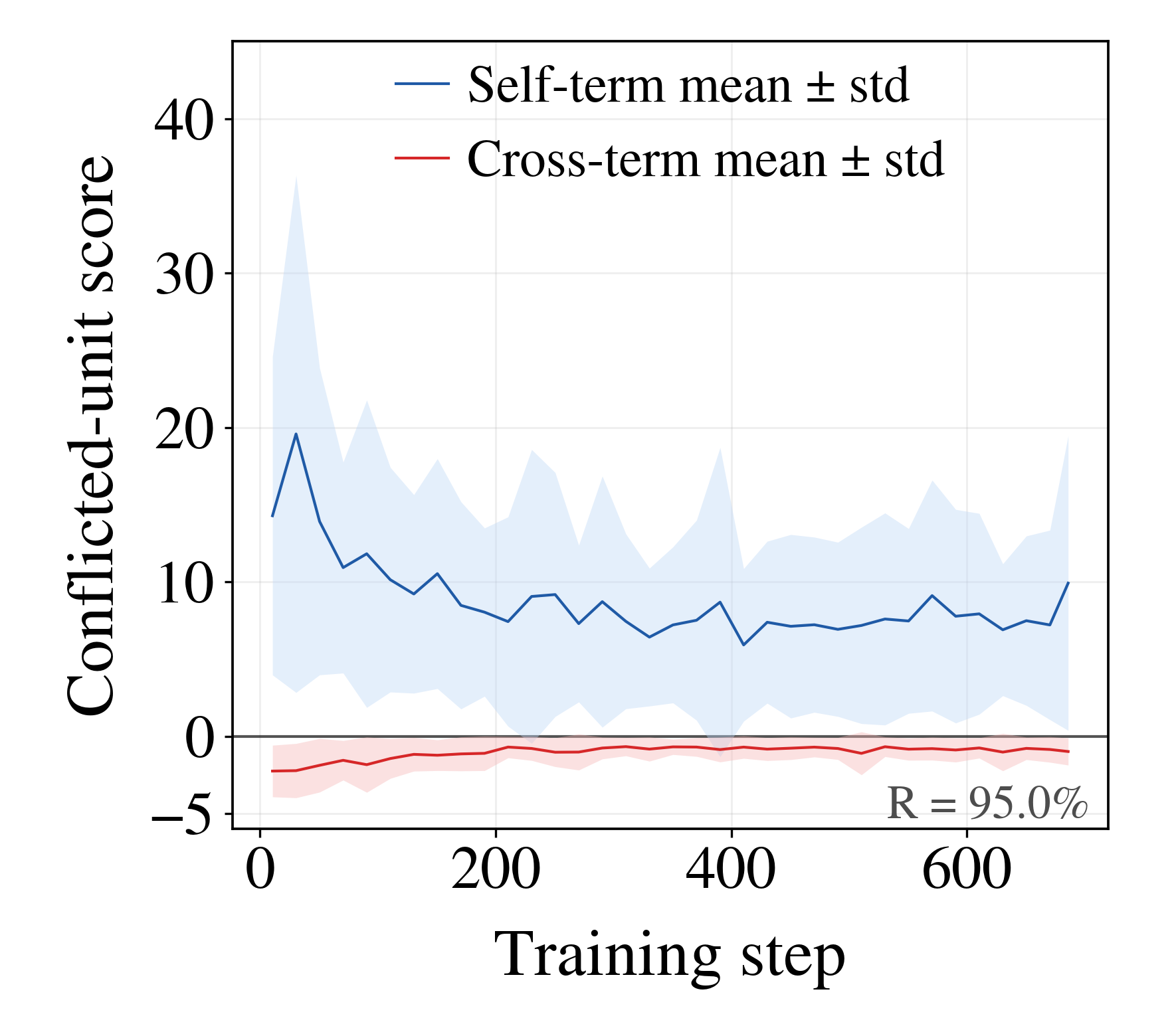}
    \vspace{-7mm}
    \captionsetup{margin={6mm,0mm}}
    \caption{Self-protected conflicts}
    \label{fig:gsm8k_self_protected_conflict}
\end{subfigure}

\vspace{-0.5mm}

\begin{subfigure}[t]{0.33\textwidth}
    \centering
    \includegraphics[width=\linewidth]{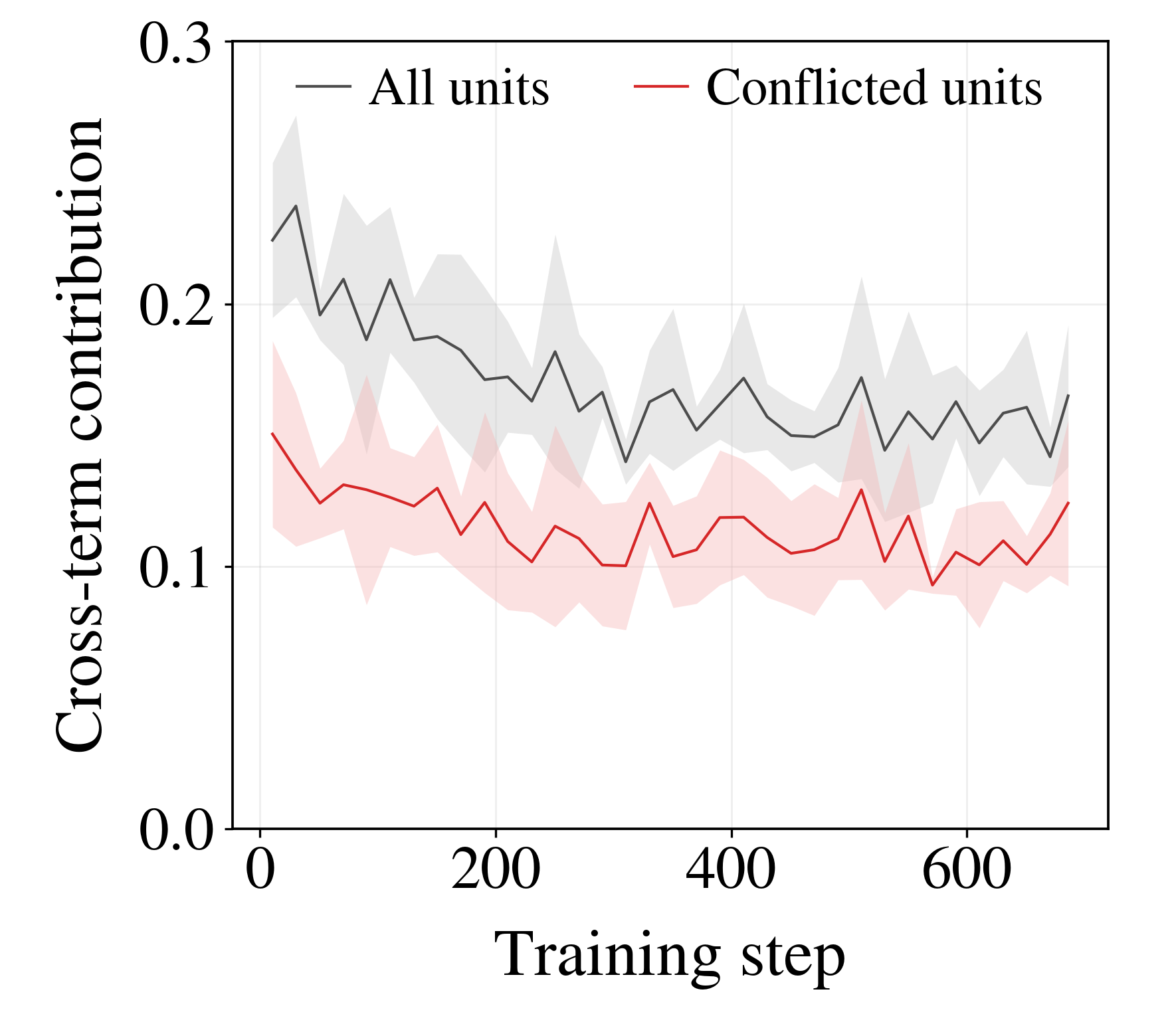}
    \vspace{-7mm}
    \captionsetup{margin={5mm,0mm}}
    \caption{Cross-term contribution}
    \label{fig:gsm8k_cross_contribution}
\end{subfigure}
\hfill
\begin{subfigure}[t]{0.33\textwidth}
    \centering
    \includegraphics[width=\linewidth]{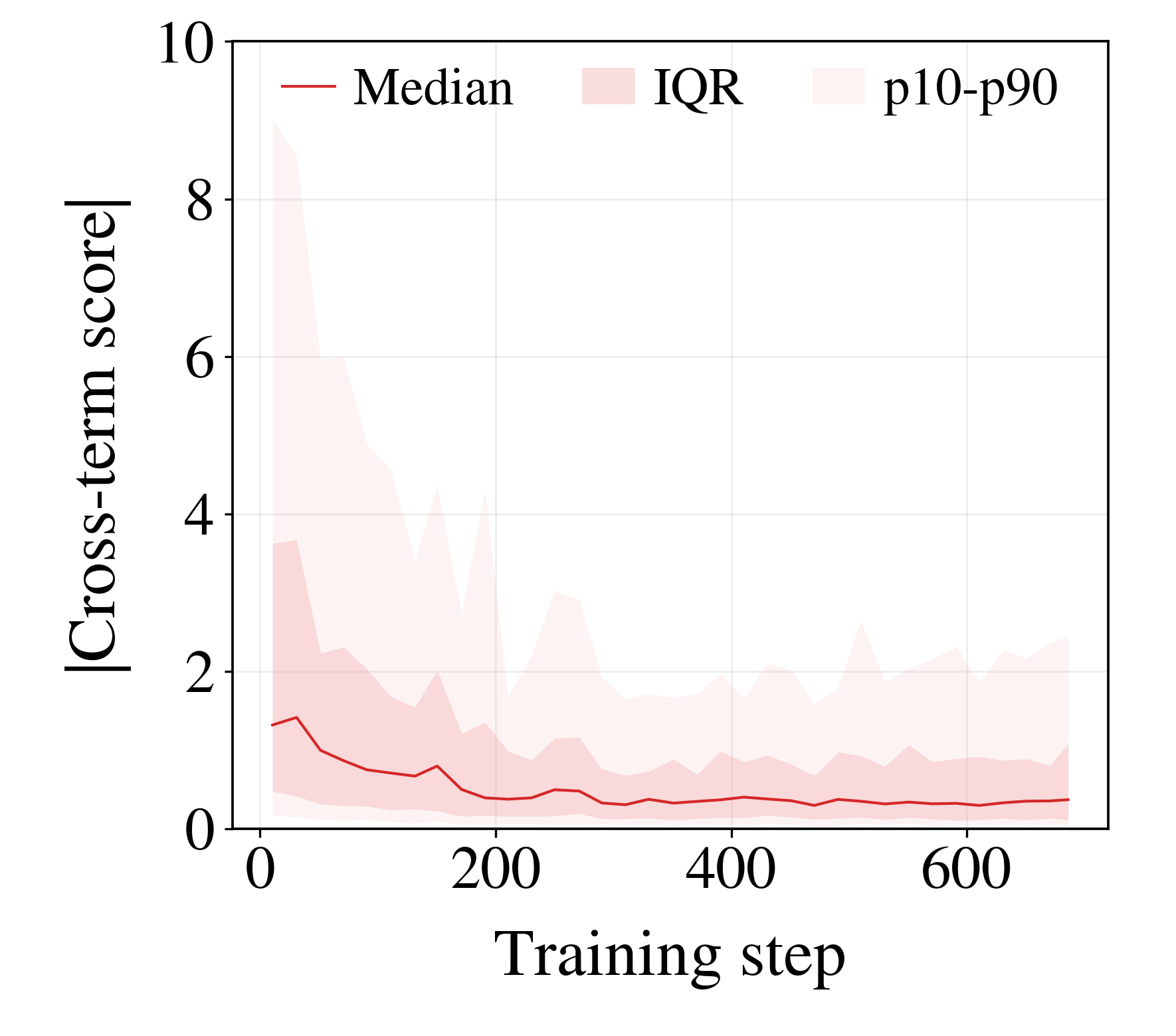}
    \vspace{-7mm}
    \captionsetup{margin={4mm,0mm}}
    \caption{Conflicted-unit cross-term magnitude}
    \label{fig:gsm8k_weak_cross}
\end{subfigure}
\hfill
\begin{subfigure}[t]{0.33\textwidth}
    \centering
    \includegraphics[width=\linewidth]{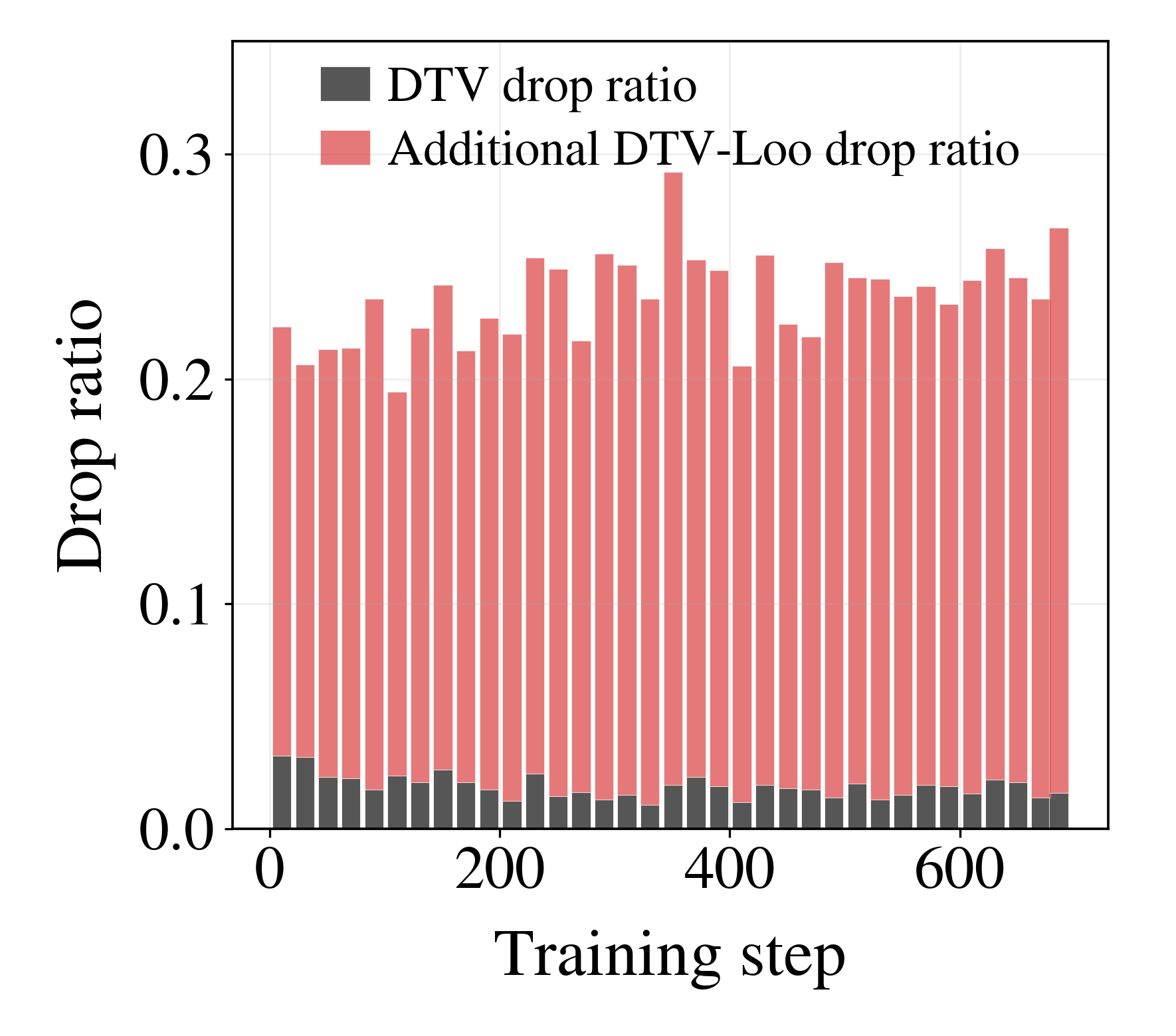}
    \vspace{-7mm}
    \captionsetup{margin={7mm,0mm}}
    \caption{Drop ratio}
    \label{fig:gsm8k_drop_ratio}
\end{subfigure}

\vspace{-1mm}

\caption{
\textbf{Empirical analysis of the self- and cross-term effect on \textit{GSM8K} under \textit{Mismatch-20\%}.} We define \emph{conflicted units} as training units for which DTV and DTV-Loo make different keep/drop decisions; $R$ denotes the fraction of units retained for visualization after trimming extreme score ranges, and IQR denotes the interquartile range (25th--75th percentiles).
(a) Decomposition of the DTV score into the self- and cross-terms.
(b) Filtering regions of DTV and DTV-Loo. 
(c) Self- and cross-term scores of conflicted units. 
(d) Cross-term contribution $|c_j|/(s_j+|c_j|)$ for all and conflicted units. 
(e) Cross-term magnitude of conflicted units.
(f) Drop ratios of DTV and the additional filtering induced by DTV-Loo.
}
\label{fig:gsm8k_analysis}
\vspace{-4mm}
\end{figure}

\begin{table*}[t]
\centering
\small
\setlength{\tabcolsep}{9.8 pt}
\caption{
\textbf{GRPO results on \textit{AIME 2024} with a 32K maximum generation length.}
$\Delta$ denotes the change relative to the shared pre-trained model. Results are from a single training run for each method, with the best result shown in bold.
}
\vspace{-2mm}
\renewcommand{\arraystretch}{1.08}
\begin{tabular}{@{}lcccccccc@{}}
\toprule
& \multicolumn{2}{c}{Pass@1}
& \multicolumn{2}{c}{Pass@16}
& \multicolumn{2}{c}{Maj@16}
& \multirow{2}{*}{Extractable $\uparrow$}
& \multirow{2}{*}{\shortstack{Acc. given\\extractable $\uparrow$}} \\
\cmidrule(lr){2-3}
\cmidrule(lr){4-5}
\cmidrule(lr){6-7}
Method
& Score $\uparrow$ & $\Delta$ $\uparrow$
& Score $\uparrow$ & $\Delta$ $\uparrow$
& Score $\uparrow$ & $\Delta$ $\uparrow$
& & \\
\midrule

Pre-trained
& 0.1063
& 0.0000
& 0.3333
& 0.0000
& 0.2333
& 0.0000
& 0.7292
& 0.1457 \\
\cmidrule(lr){1-9}

Vanilla GRPO
& 0.1063
& 0.0000
& 0.4000
& 0.0667
& 0.2333
& 0.0000
& 0.7375
& 0.1441 \\

DTV (Ours)
& 0.0917
& -0.0146
& 0.3333
& 0.0000
& 0.2667
& 0.0334
& 0.7188
& 0.1275 \\

\textbf{DTV-Loo (Ours)}
& \textbf{0.1333}
& \textbf{0.0270}
& \textbf{0.4333}
& \textbf{0.1000}
& \textbf{0.3667}
& \textbf{0.1334}
& \textbf{0.7521}
& \textbf{0.1773} \\

\bottomrule
\end{tabular}
\label{tab:grpo_aime_results}\vspace{-4mm}
\end{table*}

\textbf{Results on \textit{GSM8K}.}
Table~\ref{tab:grpo_gsm8k_results} reports the \textit{GSM8K} results under \textit{Clean} and \textit{Mismatch-20\%}. DTV achieves the best exact and partial accuracy in both settings, improving exact accuracy over the shared pre-trained model by $0.0760$ under \textit{Clean} and $0.0690$ under \textit{Mismatch-20\%}. DTV-Loo also improves over Vanilla GRPO in both settings, although DTV attains slightly higher average accuracy in both cases. In contrast, Random Filtering exhibits highly variable behavior across filtering ratios and data conditions, with the relative performance of 5\% and 10\% reversing between \textit{Clean} and \textit{Mismatch-20\%}, consistent with its sensitivity to stochastic sample removal. Reward-based Filtering is likewise unstable and does not provide reliable gains, with several settings falling below the shared pre-trained model. The two gradient-alignment baselines, LearnAlign and GradAlign, provide modest but more stable gains than Random and Reward-based Filtering across both settings, while remaining below DTV and DTV-Loo on all metrics. Under \textit{Mismatch-20\%}, their gains further diminish, with exact-accuracy improvements of only $0.0140$ and $0.0052$, respectively, compared with $0.0690$ for DTV. Overall, these results show that gradient-based adaptive filtering using only current mini-batch gradients yields stronger and more stable GRPO performance on \textit{GSM8K}. By using the batch gradient as an endogenous reference, DTV removes trajectories that conflict with the current update direction, reducing detrimental gradient interference before the model update and improving optimization performance.

Unlike the baseline methods, which rely on prescribed selection or filtering budgets, our proposed DTV and DTV-Loo make filtering decisions directly using the zero valuation threshold, allowing the effective filtering ratio to adapt dynamically over training. Notably, DTV slightly outperforms DTV-Loo across all reported metrics, suggesting that retaining the self-term can remain beneficial in this setting. To understand this difference, we revisit the self- and cross-term decomposition and examine how the self-term affects the filtering decisions of DTV and DTV-Loo on \textit{GSM8K}. We return to this distinction in the latter DPO analysis to further clarify when each variant is preferable.

Figure~\ref{fig:gsm8k_analysis} provides a closer examination of the difference between DTV and DTV-Loo, i.e., the self- and cross-term effect on \textit{GSM8K}. Figure~\ref{fig:gsm8k_analysis}(a) demonstrates that the self-term constitutes the primary component of the DTV score, whereas the cross-term remains comparatively small throughout training. This suggests that DTV may be susceptible to extreme samples that dominate the batch gradient, as the batch mean used in DTV lacks robustness to outliers. Although we do not observe this issue on \textit{GSM8K}, we revisit it in the next subsection to highlight the necessity of DTV-Loo, which mitigates the self-term effect while remaining effective at filtering detrimental samples. In addition, as implied by the decomposition in Section~\ref{subsec:dtv_loo}, removing the non-negative self-term causes DTV-Loo to additionally filter units with negative cross-terms that DTV still retains. Figure~\ref{fig:gsm8k_analysis}(b)--(c) visualize these conflicted units and their corresponding self- and cross-term scores. The remaining question is whether these additionally filtered units are sufficiently detrimental to move. Furthermore, Figure~\ref{fig:gsm8k_analysis}(d)--(e) show that conflicted units have relatively small cross-term contributions and that many of their negative cross-terms remain close to zero, indicating weak cross-unit conflicts and similar performance between DTV and DTV-Loo. Consequently, although DTV-Loo filters substantially more units through training, as shown in Figures~\ref{fig:gsm8k_analysis}(f), its performance remains close to DTV, suggesting strong data efficiency. Meanwhile, retaining these weakly conflicted units helps explain the slightly better performance of DTV in this setting.

\textbf{Scalability and Generalization Test.} To further evaluate the scalability and generalization of DTV and DTV-Loo, we consider a more complicated and large-scale \textit{AIME} challenge. Specifically, unlike the \textit{GSM8K} experiments with Gemma-3-1B-IT and LoRA, here we full-parameter fine-tune DeepSeek-R1-Distill-Qwen-1.5B using a JAX-based implementation on a dual-worker TPU v5p-16 setup, providing an additional evaluation across model backbones, fine-tuning regimes, and implementation stacks. All other experiments use a single TPU v5p-8 node, whereas \textit{AIME} requires the dual-worker setup due to HBM constraints, introducing inter-worker communication and nondeterministic task assignment at startup. A single training seed requires over 140 hours for each method; we therefore conduct one training run for each method. Given this substantial cost and the ad hoc filtering-ratio choices required by Random and Reward, we omit these two baselines and focus on Vanilla GRPO, DTV, and DTV-Loo and report the results in Table~\ref{tab:grpo_aime_results}.  

Despite this demand setting, DTV-Loo achieves the best performance across all reported metrics. Relative to the shared pre-trained model, it improves Pass@1 from $0.1063$ to $0.1333$ ($+0.0270$), Pass@16 from $0.3333$ to $0.4333$ ($+0.1000$), and Maj@16 from $0.2333$ to $0.3667$ ($+0.1334$). It also attains the highest extractable-answer rate ($0.7521$) and accuracy given an extractable answer ($0.1773$), indicating that the gains extend beyond aggregate reasoning accuracy to more reliable final-answer generation. In comparison, Vanilla GRPO and DTV show mixed improvements across the reasoning metrics, with DTV improving Maj@16 but not Pass@1 or Pass@16. 

% In this more challenging long-form reasoning setting, DTV-Loo outperforms DTV across all reported metrics. Together with the opposite ordering observed on \textit{GSM8K} and the corresponding self-term analysis, these results indicate that the empirical benefit of retaining self-term is setting dependent, consistent with Remark~\ref{rem:self_protection}.
% \jiablue{Revised.}

\begin{table*}[t]
\centering
\footnotesize
\setlength{\tabcolsep}{5.0pt}
\caption{
\textbf{DPO preference-accuracy results on \textit{UltraFeedback} under clean and mismatch settings.} Results are reported as mean $\pm$ std.\ over five seeds, with the best result in each column shown in bold. \textsuperscript{\dag}The final row reports two-sided paired $t$-test $p$-values comparing DTV-Loo with Reward 10\% across matched seeds; significant differences ($p<0.05$) are shown in bold.
}
\vspace{-2mm}
\renewcommand{\arraystretch}{1.08}
\resizebox{\textwidth}{!}{%
\begin{tabular}{lcccccc}
\toprule
& \multicolumn{2}{c}{\textit{Clean}}
& \multicolumn{2}{c}{\textit{Mismatch-20\%}}
& \multicolumn{2}{c}{\textit{Mismatch-40\%}} \\
\cmidrule(lr){2-3}
\cmidrule(lr){4-5}
\cmidrule(lr){6-7}
Method
& AUC $\uparrow$
& Acc. $\uparrow$
& AUC $\uparrow$
& Acc. $\uparrow$
& AUC $\uparrow$
& Acc. $\uparrow$ \\
\midrule

Vanilla DPO
& 0.5611 $\pm$ 0.0035
& 0.5388 $\pm$ 0.0073
& 0.5377 $\pm$ 0.0029
& 0.5014 $\pm$ 0.0044
& 0.4932 $\pm$ 0.0105
& 0.4301 $\pm$ 0.0160 \\

Random 5\%
& 0.5585 $\pm$ 0.0022
& 0.5449 $\pm$ 0.0023
& 0.5371 $\pm$ 0.0032
& 0.5013 $\pm$ 0.0152
& 0.4907 $\pm$ 0.0105
& 0.4234 $\pm$ 0.0143 \\

Random 10\%
& 0.5578 $\pm$ 0.0017
& 0.5421 $\pm$ 0.0063
& 0.5363 $\pm$ 0.0040
& 0.5015 $\pm$ 0.0100
& 0.4900 $\pm$ 0.0113
& 0.4332 $\pm$ 0.0122 \\

Reward 5\%
& 0.5620 $\pm$ 0.0016
& 0.5559 $\pm$ 0.0062
& 0.5465 $\pm$ 0.0028
& 0.5350 $\pm$ 0.0092
& 0.5083 $\pm$ 0.0089
& 0.4683 $\pm$ 0.0191 \\

Reward 10\%
& 0.5652 $\pm$ 0.0013
& 0.5647 $\pm$ 0.0063
& 0.5490 $\pm$ 0.0006
& 0.5443 $\pm$ 0.0048
& 0.5132 $\pm$ 0.0090
& 0.4857 $\pm$ 0.0147 \\

DTV (Ours)
& 0.5653 $\pm$ 0.0021
& 0.5618 $\pm$ 0.0020
& 0.5445 $\pm$ 0.0041
& 0.5236 $\pm$ 0.0068
& 0.4985 $\pm$ 0.0110
& 0.4413 $\pm$ 0.0103 \\

\textbf{DTV-Loo (Ours)}
& \textbf{0.5723 $\pm$ 0.0004}
& \textbf{0.5680 $\pm$ 0.0040}
& \textbf{0.5669 $\pm$ 0.0009}
& \textbf{0.5537 $\pm$ 0.0068}
& \textbf{0.5406 $\pm$ 0.0093}
& \textbf{0.5196 $\pm$ 0.0208} \\

\midrule
\makecell[l]{p-value\textsuperscript{\dag}}
& \textbf{0.0006}
& 0.3680
& \textbf{\ensuremath{<}\,0.0001}
& 0.1421
& \textbf{\ensuremath{<}\,0.0001}
& \textbf{0.0176} \\

\bottomrule
\end{tabular}%
}
\label{tab:dpo_main_results}\vspace{-6mm}
\end{table*}

% \vspace{8mm}

\begin{figure}[t]
    \centering
    \begin{subfigure}[t]{\textwidth}
        \centering
        \includegraphics[width=\textwidth]{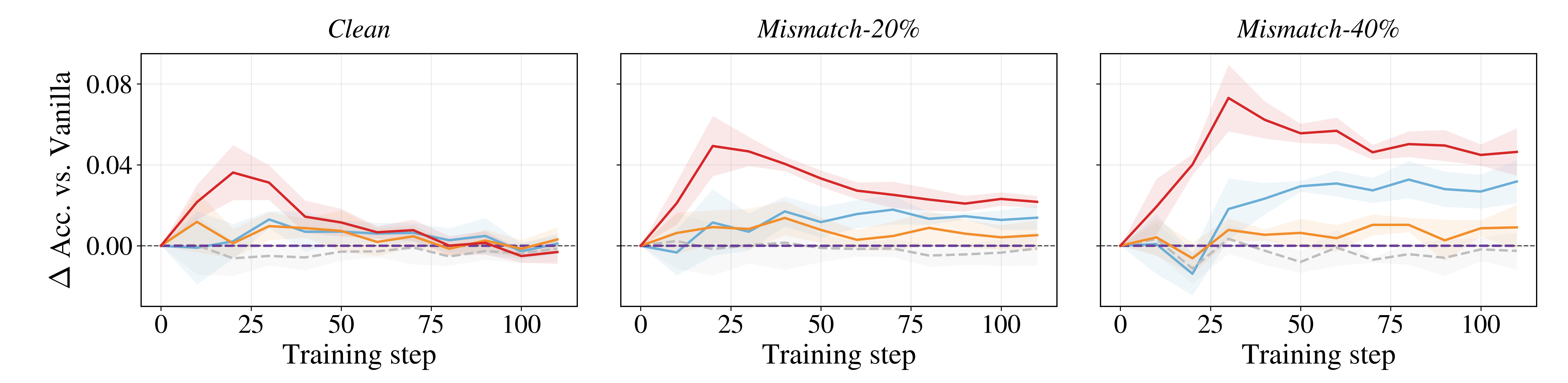}
        % \caption{Test-accuracy dynamics relative to Vanilla DPO.}
        \label{fig:dpo_delta_to_vanilla_top}
    \end{subfigure}

    \vspace{-5mm}

    \begin{subfigure}[t]{\textwidth}
        \centering
        \includegraphics[width=\textwidth]{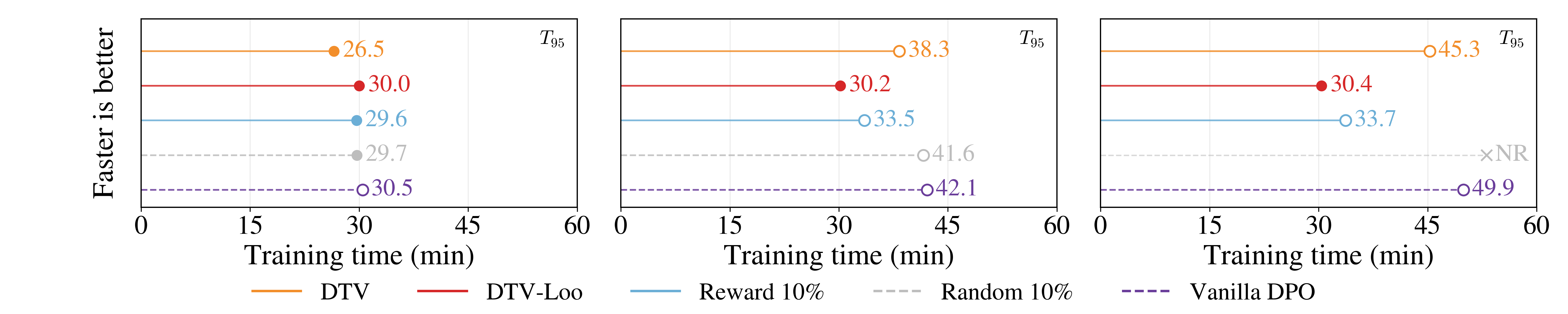}
        % \caption{Time-to-target training efficiency.}
        \label{fig:dpo_t95}
    \end{subfigure}
    \vspace{-4mm}
    \caption{ 
    \textbf{DPO validation dynamics and training efficiency on \textit{UltraFeedback}.} The top row shows the change in validation accuracy relative to Vanilla DPO throughout training under \textit{Clean}, \textit{Mismatch-20\%}, and \textit{Mismatch-40\%} conditions, averaged over five seeds; shaded regions indicate variability across seeds, and the horizontal dashed line denotes Vanilla DPO. The bottom row reports, for seed~0, the wall-clock time to reach the $T_{95}$ target defined from the five-seed best validation performance; $\times$NR denotes that the target is not reached within the training budget. DTV-Loo exhibits increasingly large and more persistent improvements as the mismatch rate increases, while also reaching the convergence target substantially faster under the mismatch settings. 
}
    \label{fig:dpo_delta_to_vanilla}\vspace{-4mm}
\end{figure}

\subsection{DTV for Pairwise Preference Valuation with DPO}
We finally turn to DPO to examine whether DTV generalizes beyond online rollout-based optimization to pairwise preference learning. In addition to the clean setting, we introduce controlled preference corruption to evaluate robustness to increasingly unreliable supervision. We further examine time-to-target efficiency and analyze the self- and cross-term decomposition underlying the self-protection effect, providing complementary views of optimization efficiency and valuation behavior.

\begin{figure}[t]
\centering
\begin{subfigure}[t]{0.33\textwidth}
    \centering
    \includegraphics[width=\linewidth]{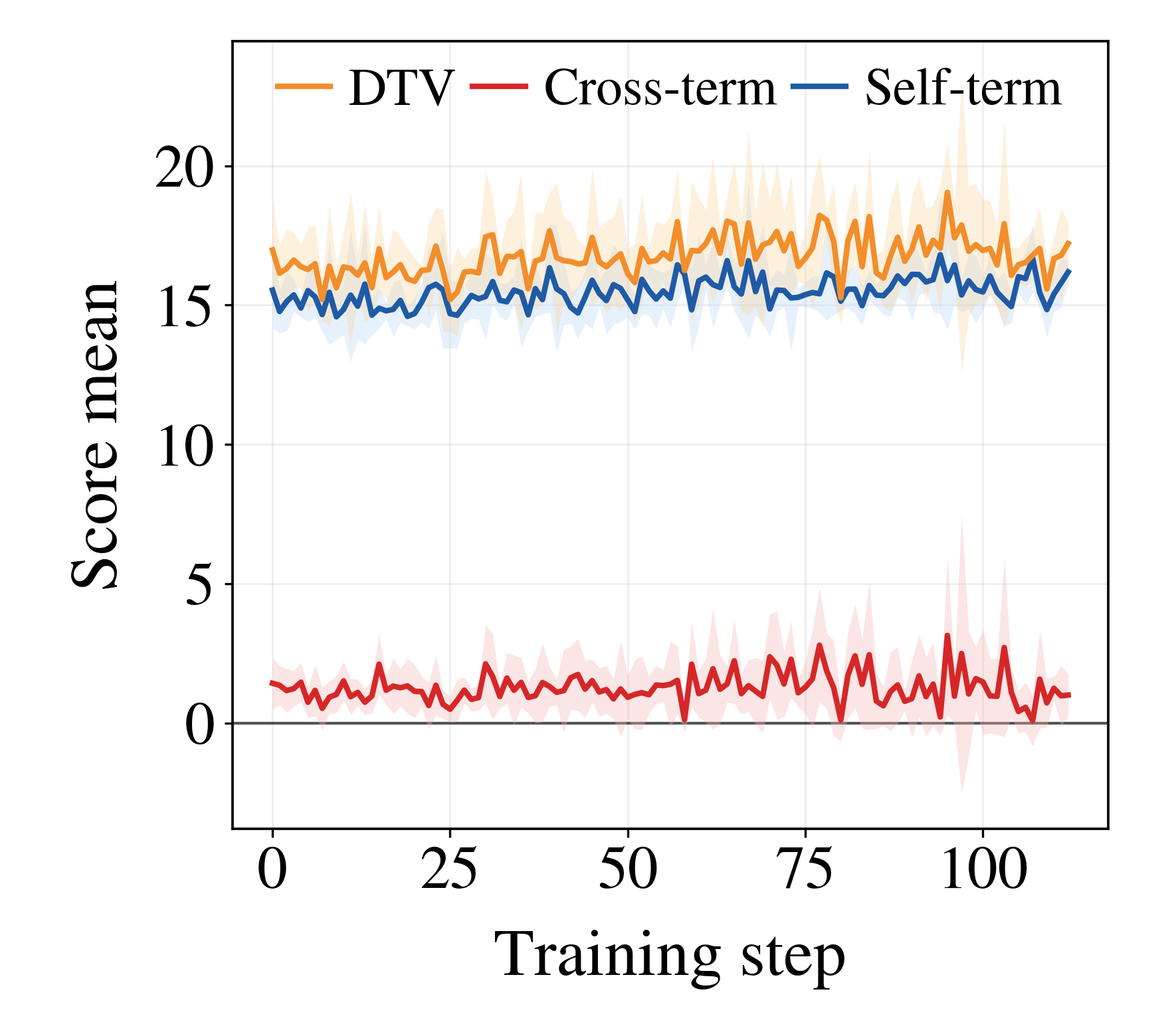}
    \vspace{-7mm}
    \captionsetup{margin={6mm,0mm}}
    \caption{Score decomposition}
    \label{fig:score_decomposition}
\end{subfigure}
\hfill
\begin{subfigure}[t]{0.33\textwidth}
    \centering
    \includegraphics[width=\linewidth]{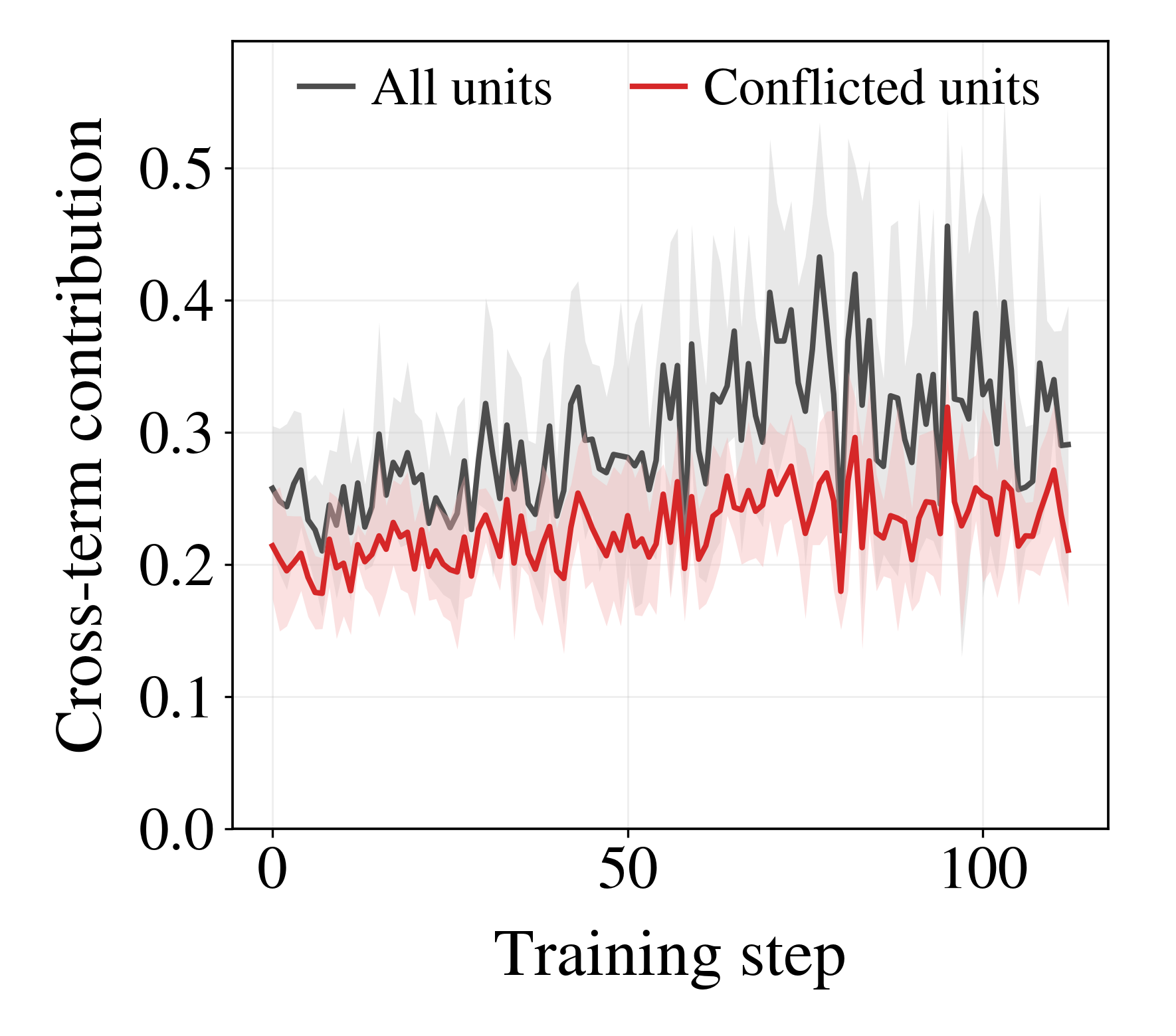}
    \vspace{-7mm}
    \captionsetup{margin={5mm,0mm}}
    \caption{Cross-term contribution}
    \label{fig:cross_term_contribution}
\end{subfigure}
\hfill
\begin{subfigure}[t]{0.33\textwidth}
    \centering
    \includegraphics[width=\linewidth]{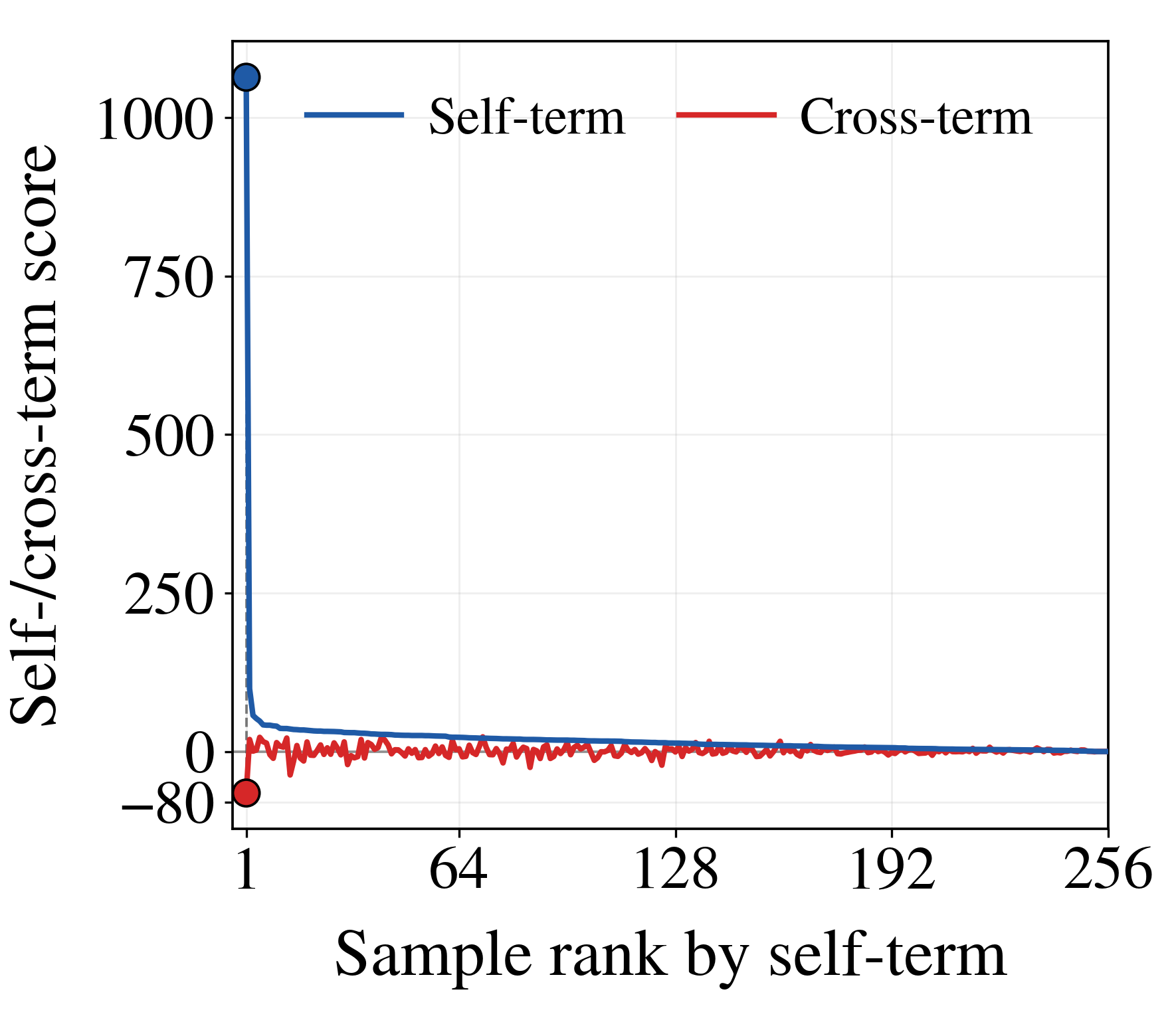}
    \vspace{-7mm}
    \captionsetup{margin={8mm,0mm}}
    \caption{Sample rank}
    \label{fig:dominated_samples}
\end{subfigure}

\caption{
\textbf{Empirical analysis of the self-protection effect on \textit{UltraFeedback} under \textit{Mismatch-20\%}.}
(a) Decomposition of the DTV Score into the self- and cross-terms.
(b) Cross-term contribution $|c_j|/(s_j+|c_j|)$ for all and conflicted units.
(c) Visualization of self- and cross-term scores across samples in a representative batch (ranked by self-term). A single dominant sample with an extreme self-term score overwhelms the cross-term interactions across the entire batch, illustrating the self-protection mechanism in standard DTV. 
}
\label{fig:self_term_analysis}\vspace{-4mm}
\end{figure}

\textbf{Experimental Setup.}
In the DPO setting, our comparison includes Vanilla DPO, Random Pair Filtering (Random), Reward-based Filtering (Reward), DTV, and DTV-Loo. Unlike online rollout-based optimization, DPO learns from prompt--response preference pairs, where misleading preferences can directly distort the optimization direction, making it a natural testbed for training-unit valuation. Vanilla DPO performs standard DPO updates without filtering. Random Pair Filtering provides a simple data-removal baseline by discarding preference pairs uniformly at random, while Reward-based Filtering prioritizes pairs according to the model's DPO reward margin. DTV and DTV-Loo instead perform adaptive preference-pair valuation: each preference pair is treated as a training unit, and pairs with negative valuation scores are filtered following the zero-threshold rule. 

Experiments are conducted on \textit{UltraFeedback}~\citep{cui2023ultrafeedback} with Qwen2.5-1.5B~\citep{qwen2.5,qwen2} using a two-stage pipeline consisting of supervised fine-tuning (SFT) followed by DPO. We consider \textit{Clean}, \textit{Mismatch-20\%}, and \textit{Mismatch-40\%}, where the latter two corrupt 20\% and 40\% of the DPO training pairs through cross-response mismatching and chosen--rejected label flipping; all held-out evaluation and test data remain clean. For both Random and Reward-based Filtering, we use fixed filtering rates of 5\% and 10\%. In contrast, DTV and DTV-Loo use the zero valuation threshold defined in Section~\ref{sec:method}, allowing the effective filtering ratio to adapt dynamically during training without tuning a prescribed filtering ratio. We report the normalized area under the held-out preference accuracy curve over the training trajectory (AUC) and the preference accuracy at the final checkpoint (Acc.). Full hyperparameter and implementation details, together with additional downstream instruction-following results, are provided in Appendix~\ref{app:dpo}.

\textbf{Main Results.} Table~\ref{tab:dpo_main_results} summarizes in-domain preference performance on the held-out test set, covering both training-time behavior and final-checkpoint accuracy. DTV-Loo achieves the best AUC across all three conditions. Compared with Reward 10\%, its AUC margin widens from $0.5723$ vs.\ $0.5652$ ($+0.0071$) under \textit{Clean} to $0.5406$ vs.\ $0.5132$ ($+0.0274$) under \textit{Mismatch-40\%}, indicating an increasingly pronounced training-time advantage as preference corruption intensifies. The advantage also extends to final-checkpoint performance, where DTV-Loo achieves the highest Acc. in all three settings, with the corresponding margin over Reward 10\% increasing from $0.5680$ vs.\ $0.5647$ ($+0.0033$) to $0.5196$ vs.\ $0.4857$ ($+0.0339$). The stronger robustness under increasing corruption is consistent with the leave-one-out motivation in Section~\ref{subsec:dtv_loo}. With a larger fraction of misleading training units, cross-unit conflicts become more consequential, while positive self-terms can mask negative alignment and cause DTV to retain these units. By removing the self-term and relying solely on cross-unit alignment, DTV-Loo suppresses this self-protection and better preserves the conflict signal, which is reflected in its stronger performance under heavier corruption. This robustness is further supported by paired $t$-tests against Reward 10\%, which show significant AUC improvements under all three conditions ($p=0.0006$, $p<0.0001$, and $p<0.0001$), while the Acc. improvement is significant under \textit{Mismatch-40\%} ($p=0.0176$). Together, these results underscore the robustness of DTV-Loo to corrupted preference supervision. DTV-Loo consistently achieves the best performance across training, outperforming both Random and Reward Filtering even when their fixed filtering rates are increased to 20\% and 40\%, which is shown in Appendix~\ref{app:dpo}.

Figure~\ref{fig:dpo_delta_to_vanilla} further characterizes these gains through test-accuracy dynamics and time-to-target efficiency. The top row shows the change in test preference accuracy relative to Vanilla DPO throughout training. Under \textit{Clean}, DTV-Loo establishes a clear early-stage advantage, although the gap gradually narrows as training proceeds. Under the mismatch settings, the advantage becomes both larger and more persistent, with the strongest separation observed under \textit{Mismatch-40\%} and substantial gains maintained through the later stages of training. Reward 10\% also improves over Vanilla DPO under mismatch, but its gains remain consistently smaller, while DTV and Random 10\% exhibit weaker or less stable improvements. %The bottom row provides a complementary view of training efficiency. Using the $T_{95}$ target defined from the best test preference accuracy across five seeds, DTV-Loo reaches the target in approximately 30 minutes on seed~0 under both mismatch settings, faster than the other methods that reach the same target. To examine whether this advantage can be explained simply by the choice of filtering ratio, we further evaluate Random and Reward Filtering with fixed rates of 20\% and 40\% under \textit{Mismatch-40\%}. As shown in Appendix~\ref{app:dpo}, DTV-Loo consistently achieves the best performance across training, outperforming both Random and Reward Filtering even when their fixed filtering rates are increased to 20\% and 40\%.

\textbf{Computational Efficiency.} Here we further demonstrate the training efficiency of our proposed DTV and DTV-Loo. As demonstrated in the bottom row of Figure~\ref{fig:dpo_delta_to_vanilla}, by dynamically filtering out harmful or conflicting trajectories to prevent counterproductive parameter updates, \mbox{DTV-Loo} enhances data efficiency, achieving target performance with substantially fewer optimization steps. Consequently, under corrupted preference settings, \mbox{DTV-Loo} reaches the target accuracy ($T_{95}$) in approximately 30 minutes—drastically reducing total wall-clock training time compared to Vanilla DPO, while Random 10\% fails to reach the target under \textit{Mismatch-40\%} within the training budget. Similar phenomena are also observed in the PPO experiments shown in Figure~\ref{fig:ppo_results}. In short, our method achieves a double win by simultaneously boosting performance and accelerating overall training.

\textbf{When to Use DTV vs. DTV-Loo.} The core distinction between DTV and DTV-Loo stems from the cross-term effect. While the previous subsection illustrated a regime where a minimal cross-term effect results in marginal performance differences between DTV and DTV-Loo (e.g., on \textit{GSM8K}), the cross-term plays a far more decisive role under other training conditions. Below, we examine scenarios where cross-term interactions substantially impact optimization (see the difference under the noisy \textit{Mismatch-20\%} setting in Table~\ref{tab:dpo_main_results}) and provide practical guidance on selecting between the two variants.

From a macro perspective, the self-term acts as a regularizing buffer that governs the strictness of trajectory filtering: retaining it favors conservatism, whereas removing it via DTV-Loo enforces strict cross-unit consistency. Figure~\ref{fig:self_term_analysis} empirically substantiates this trade-off under \textit{Mismatch-20\%}. Specifically, Figure~\ref{fig:self_term_analysis}(a) demonstrates that the DTV score is predominantly driven by the self-term throughout training. This dominance becomes critical when a training unit is negatively aligned with the batch: a strong positive self-term can easily override a negative cross-term, thereby causing DTV to retain units that DTV-Loo would otherwise filter. In contrast to the decreasing cross-term contribution observed on \textit{GSM8K}, Figure~\ref{fig:self_term_analysis}(b) shows that the cross-term contribution remains substantial and generally increases during DPO training, making cross-unit alignment more consequential in this setting. Figure~\ref{fig:self_term_analysis}(c) provides a representative case within a single mini-batch where a single sample exhibits an exceptionally large self-term, whereas the cross-terms across all samples remain negligible in magnitude. Consequently, this single outlier sample dominates the aggregate DTV valuation of the entire batch, artificially driving the score positive and masking critical cross-unit conflicts that DTV-Loo successfully identifies and removes. Moreover, the self-term violates the majority gradient assumption and disproportionately affects DTV's valuation despite its negative cross-term. By removing this unit's self-term, DTV-Loo is less susceptible to such a dominant unit and better preserves the cross-unit conflict signal. 

% --- Paragraph 2: Practical Decision Rule / Guidelines ---
These insights provide clear practical guidance on selecting between the two variants, supporting \mbox{DTV-Loo} as the default choice in practice. By eliminating self-protection, \mbox{DTV-Loo} relies strictly on cross-unit alignment, offering superior robustness, faster convergence, and significant performance gains in complex or high-noise regimes such as DPO on \textit{UltraFeedback} and GRPO on \textit{AIME}. Meanwhile, in low-conflict scenarios like GRPO on \textit{GSM8K}—where disagreements stem primarily from weak negative cross-terms—\mbox{DTV-Loo} performs comparably to standard DTV, demonstrating that strict filtering incurs minimal risk of discarding useful signals. Retaining the self-term in DTV is thus primarily beneficial when conservative filtering is specifically needed to safeguard weakly conflicting samples; otherwise, \mbox{DTV-Loo} serves as the more robust and effective default strategy.

\section{Conclusion}
In this paper, we studied trajectory valuation in reinforcement learning and proposed Dynamic Trajectory Valuation (DTV), a simple and general framework that dynamically identifies detrimental training units through mini-batch gradient alignment without requiring a predefined validation set or influence-function approximation. Building on the decomposition of DTV, we further introduced DTV-Loo to remove the self-term and analyzed the resulting self-protection effect, revealing that the utility of the self-term can depend on the underlying optimization setting. We evaluated DTV and DTV-Loo across diverse reinforcement learning and LLM post-training paradigms, including PPO, GRPO, and DPO, spanning trajectory segments, completion trajectories, and preference pairs. Extensive experiments demonstrate that dynamic trajectory valuation can improve optimization efficiency, robustness, and final performance across these settings, highlighting the practical value of training-unit-level valuation for reinforcement learning.

\section*{Acknowledgment}
We gratefully acknowledge the support of the Google TPU Builders Program and Google Tunix library, which provided support and access to the computational framework.

\newpage
\bibliographystyle{unsrtnat}   % Numeric citations in order of appearance.
\bibliography{references}

%%%%%%%%%%%%%%%%%%%%%%%%%%%%%%%%%%%%%%%%%%%%%%%%%%%%%%%%%%%%
\newpage
\appendix
\section*{Appendix}

This appendix provides additional related work, implementation details for PPO, GRPO, and DPO, and extended DPO experimental results to complement the main manuscript.

\section{Additional Related Work on on Data-Centric RL}
\label{app:relatedwork}

\begin{table}[h]
\centering
\caption{Comparison of additional requirements and applicability of gradient-based data valuation methods in reinforcement learning. The binary columns indicate whether each method uses an auxiliary valuation signal, assumes a fixed dataset, requires a validation set, uses an explicit selection budget, requires extra rollouts for valuation, or performs valuation online. The final column summarizes each method's generalizability across reinforcement learning settings.}
\label{tab:rl_valuation_comparison}
\small
\setlength{\tabcolsep}{8.6 pt}
\renewcommand{\arraystretch}{1}
% \vspace{1mm}
\begin{tabular}{lccccccc}
\toprule
Method & 
\makecell{Auxiliary\\signal} & 
\makecell{Fixed\\dataset} &
\makecell{Validation\\set} &
\makecell{Selection\\budget} & 
\makecell{Extra\\rollouts} &
% \makecell{Training set} &
\makecell{Online\\valuation} &
\makecell{Applied\\scenario} \\
\midrule

IIF~(\citeyear{hu2025snapshot})
& Yes
& \textbf{No}
& \textbf{No}
& Yes
& \textbf{No}
& \textbf{Yes}
& PPO
\\

LearnAlign~(\citeyear{li2026learnalign})
& Yes
& Yes
& \textbf{No}
& Yes
& Yes
& No
& GRPO / RLVR
\\

GradAlign~(\citeyear{yang2026gradalign})
& Yes
& \textbf{No}
& Yes
& Yes
& Yes
& \textbf{Yes}
& GRPO
\\

\textbf{DTV/DTV-Loo (Ours)}
& \textbf{No}
& \textbf{No}
& \textbf{No}
& \textbf{No}
& \textbf{No}
& \textbf{Yes}
& \textbf{General}
\\

\bottomrule
\end{tabular}
\vspace{-2mm}
\end{table}

Table~\ref{tab:rl_valuation_comparison} compares DTV with three representative gradient-based approaches and summarizes their requirements in reinforcement learning. IIF~\citep{hu2025snapshot} requires an algorithm-specific surrogate target function for valuation and a filtering parameter $p$ that determines how many negatively influential samples are removed. Its target function must also be adapted to the underlying reinforcement learning setting. IIF performs valuation online by reusing the current rollout buffer without additional rollouts. LearnAlign~\citep{li2026learnalign} relies on verifiable ground-truth feedback to estimate learnability and introduces $N$ to determine the selected subset. Its formulation assumes a fixed problem dataset with ground-truth answers, as required by the Reinforcement Learning with Verifiable Rewards (RLVR) setting, rather than experience collected purely through online environment interaction. It also requires additional rollouts to estimate the learnability. GradAlign~\citep{yang2026gradalign} requires a trusted validation set to construct the reference gradient and uses $q$ to specify the fraction of candidate problems retained. Its formulation is developed in the GRPO setting and performs valuation online by recomputing gradients under the evolving policy. GradAlign requires additional rollouts to estimate both validation and candidate gradients for selection. In contrast, our DTV and DTV-Loo perform online valuation directly from current batch gradients, without relying on a fixed dataset, auxiliary signals, or additional rollouts. Their core filtering rule applies a fixed zero threshold without an explicit filtering-rate or selection-budget hyperparameter, and applies consistently across GRPO, DPO, and PPO.

\section{Additional DTV Experimental Details for PPO}
\label{app:ppo}

This section provides additional details for the PPO experiments, including the environments, compared methods, model and optimization configurations, and evaluation protocol.

\subsection{Environments}
\label{app:ppo_environments}

We evaluate PPO on \textit{MiniGrid-Empty-8x8-v0} and \textit{MiniGrid-DoorKey-8x8-v0}~\citep{MinigridMiniworld23}, as illustrated in Figure~\ref{fig:ppo_environments}. Both are partially observable $8\times8$ sparse-reward grid worlds. Empty requires direct navigation to the goal; DoorKey requires collecting a key, unlocking a door, and then reaching the goal. \textit{MiniGrid} uses its standard shaped success reward, which decreases linearly with the fraction of the episode horizon consumed; failure receives zero. The native observation is a dictionary containing an image, direction, and mission. We apply \texttt{ImgObsWrapper}, so the policy receives only a channel-last $7\times7\times3$ symbolic image encoding object type, color, and state. The action space contains the seven standard \textit{MiniGrid} actions.

\begin{table*}[t]
\centering
\small
\caption{PPO training and evaluation configuration. The environment-step budget is the configured stopping target; collection ends after completing the final rollout round.}
\label{tab:ppo_hyperparameters}
\begin{tabular}{lcc}
\toprule
Hyperparameter & \textit{Empty-8x8} & \textit{DoorKey-8x8} \\
\midrule
Parallel environments & 16 & 16 \\
Environment-step budget & 160{,}000 & 1{,}000{,}000 \\
Rollout steps per environment & 128 & 640 \\
Transitions per rollout round & 2{,}048 & 10{,}240 \\
Optimizer mini-batch size & 64 & 256 \\
PPO epochs per round & 10 & 4 \\
Optimizer & SGD & Adam \\
Learning rate & $5\times10^{-3}$ & $3\times10^{-4}$ \\
Discount factor $\gamma$ & 0.99 & 0.99 \\
GAE trace parameter & 0.95 & 0.95 \\
PPO clipping coefficient & 0.2 & 0.2 \\
Entropy coefficient & 0 & 0.01 \\
Value-loss coefficient & 0.5 & 0.5 \\
Maximum gradient norm & 0.5 & 0.5 \\
Optimizer epsilon & -- & $10^{-8}$ \\
Advantage normalization & Yes & Yes \\
Evaluation episodes per checkpoint & 1{,}000 & 1{,}000 \\
\bottomrule
\end{tabular}
\vspace{-4mm}
\end{table*}

% \vspace{-4mm}
\textbf{Compared Methods.}  We compare DTV and DTV-Loo with Vanilla PPO and Iterative Influence-Based Filtering (IIF)~\citep{hu2025snapshot} in the PPO setting. Vanilla PPO performs standard PPO updates without filtering. Iterative Influence-Based Filtering (IIF)~\citep{hu2025snapshot} operates at the transition level. It reuses the current on-policy rollout buffer to construct a return-based impact target, computes policy-gradient similarity scores, and removes the bottom 12.5\% of records among those with negative influence. The rollout buffer is used for both PPO optimization and this local influence calculation; no separate validation rollout is sampled.

\looseness=-1DTV and DTV-Loo instead operate on trajectory units. Each environment stream is partitioned independently at episode termination and at the rollout boundary; an unfinished episode therefore contributes one observed trajectory fragment. For each unit $z_j$, the valuation gradient is computed from the mean PPO policy loss over its transitions. Using the gradient notation of Section~\ref{sec:method}, DTV compares $g_j$ with the mean gradient of all trajectory units, whereas DTV-Loo excludes $g_j$ from its reference mean. Units with negative scores are filtered. Thus, PPO uses a policy-loss score at trajectory level. The subsequent optimizer step uses a total-loss mask: all transitions in a filtered unit are removed from the complete PPO objective, including policy, value, and regularization terms. Scores are computed once per rollout round and the same mask is reused across all PPO epochs and mini-batches in that round. Filtering starts immediately on \textit{Empty-8x8} and after five unfiltered warmup rounds on \textit{DoorKey-8x8}; IIF uses the same environment-specific warmup schedule.

\textbf{Model and Optimization.} All methods use a convolutional actor--critic with two $3\times3$ convolutional layers of 16 and 32 channels, ReLU activations, and a linear projection to a shared 64-dimensional representation. Separate linear heads parameterize the categorical policy and scalar value estimate. No recurrent state is used. Within an environment, all methods share the architecture, interaction budget, and base PPO hyperparameters listed in Table~\ref{tab:ppo_hyperparameters}. The implementation uses JAX on TPU v5p-8 hardware.

\textbf{Evaluation.} At each reported checkpoint, the policy is evaluated for 1,000 episodes without parameter updates. Learning curves show the mean return across five seeds and $\pm1$ standard error of the mean. For the Worst/Best comparison in Figure~\ref{fig:ppo_results}, the 1,000 final-checkpoint returns are sorted separately for each seed. The lowest and highest 20 returns are pooled across seeds, giving up to 100 observations in each environment--method subset. In the split violin plots, thin lines show the selected range, thick lines the interquartile range, horizontal marks the median, and white diamonds the mean.

\section{Additional DTV Experimental Details for GRPO}
\label{app:grpo}

This section provides additional details for the GRPO experiments on \textit{GSM8K} and \textit{AIME}, including the datasets and corruption protocol, compared methods, model and optimization configurations, and evaluation settings.

\subsection{Datasets and Benchmarks}
\label{app:grpo_datasets}

\textbf{\textit{GSM8K}.} We use the official \textit{GSM8K} splits~\citep{cobbe2021gsm8k}. The data are independently shuffled for each run; the first 3,072 shuffled examples form 768 batches of four prompts. The first 691 batches (2,764 prompt occurrences) are used for training and the remaining 77 batches (308 prompts) for periodic clean evaluation. Thus, one run performs 691 updates over this selected partition. The official 1,319-example test split is used only for final evaluation. Prompts request reasoning inside \texttt{<reasoning>} tags and one numerical answer inside \texttt{<answer>} tags; targets are parsed from the substring following the \textit{GSM8K} \texttt{\#\#\#\#} delimiter.

\begin{table*}[t]
\centering
\footnotesize
\setlength{\tabcolsep}{6pt}
\caption{GRPO training and evaluation configurations on \textit{GSM8K} and \textit{AIME 2024}}
\label{tab:grpo_hyperparameters}
\begin{tabularx}{\textwidth}{
    @{}
    >{\raggedright\arraybackslash}p{0.34\textwidth}
    >{\raggedright\arraybackslash}X
    >{\raggedright\arraybackslash}X
    @{}
}
\toprule
Hyperparameter & \textit{GSM8K} & \textit{AIME} \\
\midrule
Prompts per update
    & 4 
    & 128 \\
Completions per prompt
    & 4 
    & 8 \\
Train micro-batch size
    & --
    & 2 \\
Maximum prompt / response length
    & 256 / 768
    & 1,024 / 8,192 \\
Optimizer / peak learning rate
    & AdamW / $1\times10^{-6}$
    & AdamW / $1\times10^{-6}$ \\
Learning-rate schedule
    & 69-step warmup, cosine decay
    & cosine decay, no warmup \\
Optimizer updates
    & 691
    & 314 \\
KL coefficient
    & 0.08
    & 0.001 \\
Clipping width
    & 0.20 / 0.20
    & 0.20 / 0.28 \\
Adam coefficients / weight decay
    & $(0.9,0.99)$ / 0.1
    & $(0.9,0.99)$ / 0.1
    \\
Loss aggregation
    & sequence-mean-token-mean
    & sequence-mean-token-mean \\
\midrule
Final evaluation
    & 1,319 problems, greedy decoding
    & 30 problems, 16 responses each \\
Evaluation temperature / top-$p$
    & greedy ($\mathrm{top}\text{-}k=1$)
    & 0.6 / 0.95 \\
Maximum evaluation generation
    & 768
    & 32,768 \\
\bottomrule
\end{tabularx}
\vspace{-4mm}
\end{table*}

In \textit{Mismatch-20\%}, a stable SHA256 hash of the prompt, corruption schema, and experiment seed independently selects groups at an approximately 20\% rate. The four observed rewards in each selected group are reassigned in reverse rank order. This preserves the reward multiset, mean, and standard deviation while corrupting the completion--reward association used to compute advantages. Tied rewards can make a selected group only partially changed, or unchanged when all four rewards are equal. Periodic and final evaluation data remain clean. We restrict the corruption study to \textit{Mismatch-20\%}, since extending the same rank-reversal perturbation to 40\% of prompt groups would substantially distort the within-group relative signal that directly drives the GRPO update.

\textbf{\textit{AIME}.} We train on the \textit{DeepScaleR-Preview-Dataset} and evaluate on all 30 \textit{AIME 2024} problems~\citep{huggingfaceh4_2024_aime,deepscaler2025}. After preprocessing, 40,300 training prompts are valid and 40,192 are selected after a fixed shuffle. With 128 prompts per update, this partition yields exactly 314 optimizer steps. \textit{AIME} uses binary mathematical correctness as the group reward and one training seed because of the cost of 8K-token training rollouts and 32K-token evaluation generations.

\subsection{Experimental Setup}
\label{app:grpo_experimental_setup}

\textbf{Compared Methods.} 
We compare Vanilla GRPO, Random Filtering, Reward-based Filtering, LearnAlign~\citep{li2026learnalign}, GradAlign~\citep{yang2026gradalign}, DTV, and DTV-Loo. Random Filtering, Reward-based Filtering, DTV, and DTV-Loo make completion-level filtering decisions within each prompt group, whereas LearnAlign and GradAlign perform selection at the prompt level. Vanilla GRPO retains the complete group. Random and Reward-based Filtering use prescribed expected removal rates: Random selects completions independently of reward, whereas Reward-based Filtering removes completions with the lowest observed group-relative advantages. LearnAlign performs a prompt-level selection stage before the main RLVR training, after warm-up, ranking prompts using learnability-weighted gradient alignment computed from eight rollouts per prompt and retaining a predefined top-$N$ subset; in our experiments, $N$ is set to retain 50\% of the prompts. GradAlign instead performs online prompt-level selection by ranking prompts according to their alignment with a validation gradient estimated from a small validation set using eight rollouts per validation prompt and retaining a fixed fraction (25\% in this setting). We carefully reimplement both methods in JAX for TPU execution, following their original formulations as closely as possible and choosing the selection ratios with reference to the configurations used in the original studies. DTV and DTV-Loo instead perform adaptive completion-level filtering using the current mini-batch policy gradients as an endogenous reference and the zero valuation threshold, without prescribing an overall filtering rate. On \textit{GSM8K}, we evaluate all methods, with Random and Reward-based Filtering using expected removal rates of 5\% and 10\%; on the substantially more expensive \textit{AIME} setting, we compare the three core methods, Vanilla GRPO, DTV, and DTV-Loo.

\looseness=-1\textbf{Models and Optimization.} \textit{GSM8K}. The actor and frozen reference policy are initialized from Gemma-3-1B-IT~\citep{gemma_2025}. The actor uses LoRA with rank 64 and scaling 64 on \texttt{q\_einsum}, \texttt{kv\_einsum}, \texttt{attn\_vec\_einsum}, \texttt{gate\_proj}, \texttt{down\_proj}, and \texttt{up\_proj}; base parameters remain frozen. DTV inner products therefore cover all trainable LoRA tensors.

\looseness-1Each update samples four completions for each of four prompts, with one generated completion trajectory treated as one valuation unit. Each completion is scored using the gradient of its policy-only GRPO loss, excluding KL. DTV uses all four completion gradients to construct the group reference, whereas DTV-Loo uses the other three gradients for the evaluated completion. Advantages are computed before filtering and are not recomputed after selection. After thresholding at zero, the selected completion mask is applied as a total-loss mask to the full policy-plus-KL GRPO update.

\begin{table*}[t]
\centering
\footnotesize
\setlength{\tabcolsep}{7pt}
\caption{SFT and DPO training configurations}
\label{tab:dpo_training_hyperparameters}
\begin{tabularx}{\textwidth}{
    @{}
    >{\raggedright\arraybackslash}p{0.34\textwidth}
    >{\raggedright\arraybackslash}X
    >{\raggedright\arraybackslash}X
    @{}
}
\toprule
Hyperparameter & SFT & DPO \\
\midrule
Fine-tuning mode
    & Full parameter
    & LoRA ($r=64$, scaling $=64$) \\
Train / evaluation batch size
    & 2 / 2
    & 8 / 8 \\
Gradient accumulation steps
    & 4
    & 32 \\
Maximum length
    & Target: 768
    & Prompt / response: 512 / 512 \\
Optimizer / peak learning rate
    & AdamW / $1\times10^{-5}$
    & AdamW / $1\times10^{-6}$ \\
Warmup / decay steps
    & 100 / 1,500
    & 10 / 115 \\
Weight decay
    & 0.05
    & 0.1 \\
Maximum gradient norm
    & 1.0
    & -- \\
DPO coefficient $\beta$ / label smoothing
    & --
    & 0.01 / 0 \\
Evaluation / save interval
    & 100 / 100 steps
    & 10 / 3 steps \\
Optimizer updates
    & 1,493
    & 115 \\
\bottomrule
\end{tabularx}
\vspace{-4mm}
\end{table*}

The shaped training reward combines: (i) a format score of $+1$ for the requested reasoning/answer structure and $-1$ otherwise; (ii) a numerical score of $+4$ for exact correctness and $+3,+2,+1,$ or $+0.25$ for relative errors within 1\%, 5\%, 10\%, or 25\%, with $-1$ for larger or unparseable errors; and (iii) an additional $+1.5$ for an exactly correct number after the \texttt{<answer>} tag. Using the notation in the main-text GRPO formulation, advantages are computed from the complete group as the reward minus the group mean, divided by the group sample standard deviation plus $10^{-4}$. Filtering does not recompute these statistics.  Because 5\% and 10\% are not integral for groups of four, Random and Reward-based Filtering remove one completion with probability 0.2 and 0.4, respectively. For \textit{GSM8K} DTV-Loo only, a 25\% per-group minimum retains at least one finite-score completion; an all-negative group restores its highest-scoring completion. DTV has no such minimum.

\looseness-1\textit{AIME 2024.}
We full-parameter fine-tune DeepSeek-R1-Distill-Qwen-1.5B on the selected \textit{DeepScaleR} prompts. Each update contains 128 prompts and eight completions per prompt, with one completion trajectory treated as one valuation unit. Each completion is scored from its policy-only loss gradient within the eight-completion prompt group: DTV uses all eight gradients in the reference, while DTV-Loo uses the other seven. After zero-threshold selection, the completion mask is applied as a total-loss mask to the full policy-plus-KL GRPO update. A shared mask removes groups whose eight advantages are all zero. \textit{AIME} DTV-Loo has no minimum-retention rule, so an all-negative finite group may retain no completion.

\textbf{Evaluations.}
\textit{GSM8K.} 
We use five training seeds. Methods matched by seed share data order, rollout randomness, mismatch selection, and stochastic filtering randomness. Periodic clean evaluation every 32 steps is not used for early stopping; final metrics use step 691 and the official 1,319-example test split. Exact accuracy requires a correct parsed number, while partial accuracy accepts a prediction-to-target ratio in $[0.9,1.1]$. Means and sample standard deviations are reported over independently trained seeds. Training uses JAX on one TPU v5p-8 worker with a $(4,1)$ mesh over fully sharded data parallel (FSDP) and tensor parallel (TP) axes.

\textit{AIME.}
All final checkpoints use one fixed engine-level evaluation seed and matched prompts, tokenizer, and decoding parameters. For each of the 30 problems, we sample 16 responses; these are consecutive samples from the seeded generation stream rather than independent training seeds. If a response contains \texttt{</think>}, only the subsequent answer portion is used before boxed-answer extraction. Grading applies the MathD, SymPy, and special-handling checks in \texttt{tunix.utils.math\_eval\_metrics}; a valid \textit{AIME} answer must normalize to an integer in $[0,999]$. We report Pass@1, Pass@16, Maj@16, extractability, and accuracy conditioned on extraction.

Training uses JAX on a dual-worker TPU v5p-16 setup. Vanilla GRPO requires approximately 120--144 hours, while DTV and DTV-Loo require approximately 168--192 hours because of per-completion valuation. One 32K checkpoint evaluation takes approximately eight generation-hours; all four checkpoints require 32.1 hours sequentially, excluding initialization and restoration. This substantial cost motivates the single-training-seed protocol.

\section{Additional DTV Experimental Details and Results for DPO}
\label{app:dpo}

This section provides additional details for the DPO experiments, including the dataset and mismatch construction, compared methods, model and optimization configurations, and evaluation protocol. We further report downstream instruction-following results and additional analyses of training dynamics and filtering rates.

\subsection{Dataset and Mismatch Construction}
\label{app:dpo_data}

\label{app:dpo_downstream}
\begin{table*}[t]
\small
\centering
\footnotesize
\setlength{\tabcolsep}{12 pt}
\caption{\textit{Clean} downstream instruction-following performance of DPO methods. Results are mean $\pm$ sample standard deviation over five training seeds. These metrics are not used for training, filtering, or checkpoint selection; the best result in each condition and column is shown in bold.}
\label{tab:dpo_downstream_results}
\begin{tabular}{@{}llccc@{}}
\toprule
Condition & Method & \textit{LiveBench-IF} $\uparrow$ & \textit{RB2} Precise IF $\uparrow$ & \textit{IFBench} P-Strict $\uparrow$ \\
\midrule
\multirow{7}{*}{\textit{Clean}}
& Vanilla DPO & 0.2418$\pm$0.0035 & \textbf{0.4938$\pm$0.0285} & 0.1280$\pm$0.0045 \\
& Random 5\% & 0.2413$\pm$0.0046 & 0.4763$\pm$0.0294 & 0.1227$\pm$0.0090 \\
& Random 10\% & \textbf{0.2436$\pm$0.0049} & 0.4813$\pm$0.0442 & \textbf{0.1307$\pm$0.0068} \\
& Reward 5\% & 0.2388$\pm$0.0064 & 0.4800$\pm$0.0203 & 0.1253$\pm$0.0016 \\
& Reward 10\% & 0.2370$\pm$0.0055 & 0.4613$\pm$0.0191 & 0.1247$\pm$0.0050 \\
& DTV (Ours) & 0.2430$\pm$0.0034 & \textbf{0.4938$\pm$0.0198} & 0.1247$\pm$0.0045 \\
& DTV-Loo (Ours) & 0.2415$\pm$0.0081 & 0.4238$\pm$0.0195 & 0.1293$\pm$0.0065 \\
\midrule
\multirow{7}{*}{\textit{Mismatch-20\%}}
& Vanilla DPO & \textbf{0.2454$\pm$0.0042} & 0.4950$\pm$0.0278 & 0.1233$\pm$0.0047 \\
& Random 5\% & 0.2404$\pm$0.0052 & 0.4875$\pm$0.0088 & 0.1273$\pm$0.0061 \\
& Random 10\% & 0.2434$\pm$0.0041 & \textbf{0.5025$\pm$0.0350} & 0.1267$\pm$0.0060 \\
& Reward 5\% & 0.2421$\pm$0.0025 & 0.4775$\pm$0.0196 & 0.1253$\pm$0.0081 \\
& Reward 10\% & 0.2429$\pm$0.0067 & 0.4450$\pm$0.0254 & 0.1233$\pm$0.0073 \\
& DTV (Ours) & 0.2400$\pm$0.0051 & 0.4863$\pm$0.0327 & 0.1287$\pm$0.0058 \\
& DTV-Loo (Ours) & 0.2413$\pm$0.0084 & 0.4613$\pm$0.0320 & \textbf{0.1293$\pm$0.0039} \\
\midrule
\multirow{7}{*}{\textit{Mismatch-40\%}}
& Vanilla DPO & 0.2434$\pm$0.0056 & \textbf{0.5638$\pm$0.0228} & 0.1247$\pm$0.0034 \\
& Random 5\% & 0.2492$\pm$0.0054 & 0.5300$\pm$0.0174 & 0.1260$\pm$0.0049 \\
& Random 10\% & \textbf{0.2497$\pm$0.0057} & 0.5600$\pm$0.0370 & 0.1233$\pm$0.0084 \\
& Reward 5\% & 0.2425$\pm$0.0053 & 0.5213$\pm$0.0122 & 0.1193$\pm$0.0080 \\
& Reward 10\% & 0.2448$\pm$0.0055 & 0.5050$\pm$0.0327 & 0.1193$\pm$0.0049 \\
& DTV (Ours) & 0.2445$\pm$0.0084 & 0.5588$\pm$0.0196 & \textbf{0.1267$\pm$0.0037} \\
& DTV-Loo (Ours) & 0.2357$\pm$0.0063 & 0.4563$\pm$0.0285 & 0.1233$\pm$0.0056 \\
\bottomrule
\end{tabular}
\end{table*}

We use \texttt{HuggingFaceH4/ultrafeedback\_binarized}, derived from \textit{UltraFeedback}~\citep{cui2023ultrafeedback}. Prompt-level partitions are constructed with a fixed seed. From \texttt{train\_prefs}, 25\% of prompts are assigned to SFT and 75\% to DPO; 10\% of each stage-specific partition is held out for clean periodic evaluation. SFT prompts therefore do not reappear in DPO training, and held-out prompts never contribute to training, valuation, or optimization. \texttt{test\_prefs} is used only for final-checkpoint evaluation.

\looseness-1Mismatch modifies only the DPO training partition. For the \textit{Mismatch-20\%} and \textit{Mismatch-40\%} conditions, we sample without replacement exactly the floor of the requested fraction of training pairs using the experiment seed. Donors are restricted to this selected subset, and a shuffled cyclic map prevents self-mapping. The \texttt{cross\_response\_flip} operation keeps the target prompt, replaces its preferred response with a dispreferred response from a selected non-self donor, and replaces its dispreferred response with a preferred response from another selected non-self donor. When at least three pairs are selected, the two response fields use opposite cyclic shifts; with two pairs, they necessarily share the only non-self donor. This single operation jointly introduces cross-response mismatch and preference-label inversion. Methods matched by seed use identical corrupted pairs; different seeds corrupt independently, and all evaluation data remain clean.

\subsection{Experimental Setup}
\label{app:dpo_experimental_setup}

\textbf{Compared Methods.}
We compare Vanilla DPO, Random Pair Filtering, Reward-based Filtering, DTV, and DTV-Loo. Vanilla DPO uses every preference pair. Random and Reward-based Filtering remove 5\% or 10\% of the pairs in a gradient-accumulation window; Random samples uniformly, whereas Reward-based Filtering removes pairs with the lowest DPO implicit reward margins. These baselines therefore require a prescribed filtering rate.

DTV and DTV-Loo instead treat each prompt--response preference pair as one training unit. Unlike PPO and GRPO, their score gradient is computed from the complete per-pair DPO loss rather than a separate policy-only surrogate. Gradients are taken with respect to every trainable LoRA tensor across the complete accumulation window. DTV compares each pair's gradient with the average gradient over the complete window, whereas DTV-Loo forms the reference average from the remaining pairs. Non-negative finite scores are retained. The optimizer then applies a total-loss mask to the same complete DPO objective and normalizes the final loss and gradient by the retained-pair count. There is no fixed minimum retained fraction; if an entire window would otherwise be empty, the implementation restores its finite pairs to keep the update well defined.

\textbf{Model and Optimization.}
We start from Qwen2.5-1.5B~\citep{qwen2.5,qwen2}. The base model is first full-parameter fine-tuned on the SFT partition and exported. The iterator ends after 1,493 updates, slightly before the configured 1,500-step maximum; this is dataset exhaustion rather than early stopping. Every DPO actor and frozen reference model is initialized from this same SFT export.

\begin{figure}[t]
\centering
\includegraphics[width=\textwidth]{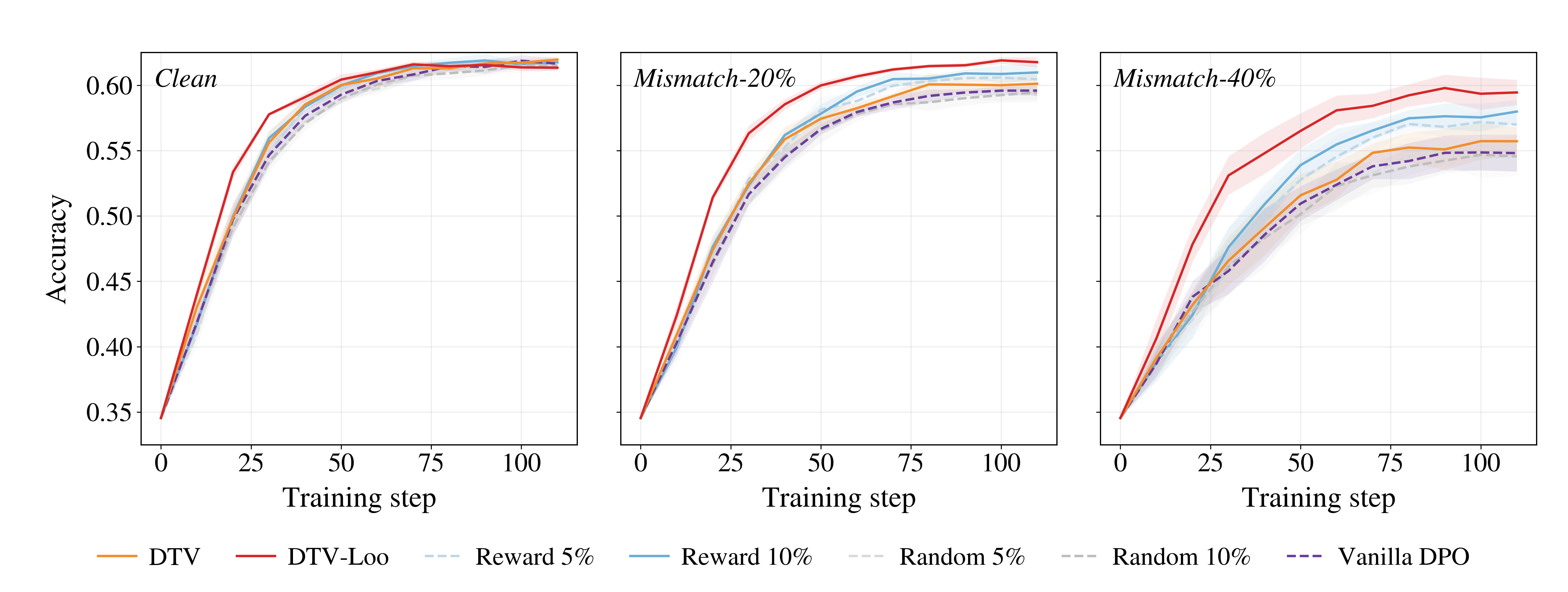}
\caption{Absolute preference-accuracy trajectories on \textit{UltraFeedback} under \textit{Clean}, \textit{Mismatch-20\%}, and \textit{Mismatch-40\%} training conditions. Curves show the five-seed mean and shaded regions show $\pm1$ sample standard deviation.}
\label{fig:dpo_absolute_accuracy}
\vspace{-4mm}
\end{figure}

The DPO actor uses LoRA with rank 64 and scaling 64 on \texttt{q\_proj}, \texttt{k\_proj}, \texttt{v\_proj}, \texttt{o\_proj}, \texttt{gate\_proj}, \texttt{up\_proj}, and \texttt{down\_proj}. Actor and reference models use a $(2,2)$ mesh over fully sharded data parallel (FSDP) and tensor parallel (TP) axes. Tables~\ref{tab:dpo_training_hyperparameters} gives the complete optimization settings.

% \begin{table*}[t]
% \small
% \centering
% \begin{minipage}[t]{0.48\textwidth}
% \vspace{0pt}
% \centering
% \footnotesize
% \caption{SFT configuration}
% \label{tab:dpo_sft_hyperparameters}
% \begin{tabularx}{\linewidth}{@{}>{\raggedright\arraybackslash}Xl@{}}
% \toprule
% Hyperparameter & Value \\
% \midrule
% Fine-tuning mode & full parameter \\
% Train / evaluation batch size & 2 / 2 \\
% Gradient accumulation & 4 \\
% Maximum target length & 768 \\
% Optimizer; peak learning rate & AdamW; $1\times10^{-5}$ \\
% Warmup / decay steps & 100 / 1,500 \\
% Weight decay; maximum gradient norm & 0.05; 1.0 \\
% Evaluation / save interval & 100 / 100 steps \\
% Realized updates & 1,493 \\
% \bottomrule
% \end{tabularx}
% \end{minipage}
% \hfill
% \begin{minipage}[t]{0.48\textwidth}
% \vspace{0pt}
% \small
% \centering
% \footnotesize
% \caption{DPO training configuration}
% \label{tab:dpo_hyperparameters}
% \begin{tabularx}{\linewidth}{@{}>{\raggedright\arraybackslash}Xl@{}}
% \toprule
% Hyperparameter & Value \\
% \midrule
% Train / evaluation batch size & 8 / 8 \\
% Gradient accumulation steps & 32 \\
% Maximum prompt / response length & 512 / 512 \\
% DPO coefficient $\beta$; label smoothing & 0.01; 0 \\
% Optimizer; peak learning rate & AdamW; $1\times10^{-6}$ \\
% Adam coefficients; weight decay & $(0.9,0.99)$; 0.1 \\
% Warmup / decay steps & 10 / 115 \\
% Evaluation / save interval & 10 / 3 steps \\
% Optimizer updates & 115 \\
% \bottomrule
% \end{tabularx}
% \end{minipage}
% \end{table*}

\textbf{Evaluation.}
All methods use the final step-115 checkpoint; periodic evaluation is not used for checkpoint selection. We train five matched seeds, with run, data-shuffle, and curation seeds matched across methods. AUC is the normalized area under the clean held-out preference-accuracy curve over training, and Acc. is the final-checkpoint preference accuracy. Held-out pairs never enter DTV's reference gradient. Means and sample standard deviations are reported over training seeds, and the paired two-sided tests in Table~\ref{tab:dpo_main_results} compare DTV-Loo with Reward 10\% on matched seeds.

\subsection{Additional Experimental Results}
\label{app:dpo_additional_results}

\subsubsection{Downstream Instruction-Following Results}

\looseness=-1We evaluate final checkpoints on three clean external instruction-following benchmarks. \textit{LiveBench-IF}~\citep{white2025livebench} evaluates on the benchmark's test split and deterministic generation with a fixed seed, top-$k$ 50, top-$p$ 0.95, maximum prompt length 4,096, maximum response length 1,024, and batch size eight. \textit{RewardBench 2}~\citep{malik2026rewardbench} evaluates on its official test split; candidates are scored by the actor's DPO implicit reward relative to the frozen SFT reference, and we report the Precise IF subset. \textit{IFBench}~\citep{pyatkin2026generalizing} uses the official prepared assets and reports prompt-level strict accuracy (P-Strict), with a fixed seed, top-$k$ 50, top-$p$ 0.95, maximum prompt length 4,096, maximum response length 1,024, and batch size eight. Table~\ref{tab:dpo_downstream_results} reports mean and sample standard deviation over five independently trained checkpoints. Downstream instruction-following results are mixed: DTV remains broadly competitive, while the stronger in-domain gains of DTV-Loo do not uniformly transfer across all external metrics.

\subsubsection{Training Curves and Filtering-Rate Ablation}
\label{app:dpo_additional_results}

\begin{figure}[t]
\centering
\includegraphics[width=0.6\textwidth]{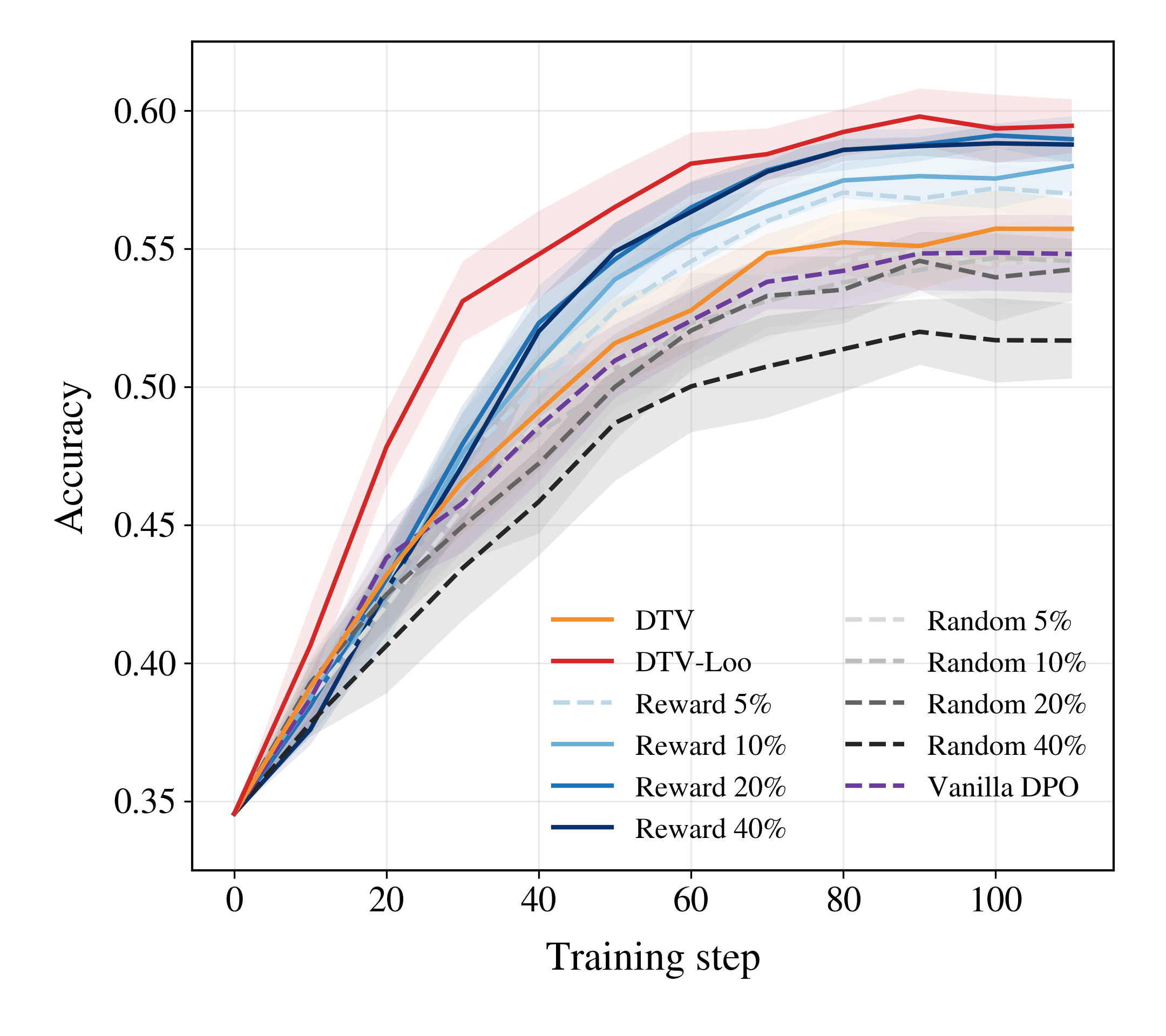}
\caption{Fixed-filtering-rate ablation under \textit{Mismatch-40\%}. DTV and DTV-Loo are compared with Random and Reward-based Filtering at 5\%, 10\%, 20\%, and 40\%. Curves show the five-seed mean, with shading denoting $\pm1$ sample standard deviation.}
\label{fig:dpo_filter_rate_ablation}
\vspace{-4mm}
\end{figure}

Figure~\ref{fig:dpo_absolute_accuracy} complements the delta-to-Vanilla view in the main text by showing the absolute preference-accuracy trajectories. Each curve is averaged over five training seeds, and shading denotes one sample standard deviation. Periodic evaluation is always clean, including for models trained under mismatch. Under \textit{Clean}, most methods converge to similar final accuracy, whereas the performance differences become increasingly pronounced as the mismatch level increases. In particular, DTV-Loo maintains faster and stronger improvement under both mismatch settings, with the largest separation observed under \textit{Mismatch-40\%}.

The main comparison uses fixed filtering rates of 5\% and 10\% for Random and Reward-based Filtering. To test whether DTV-Loo's advantage under severe mismatch is explained only by removing more pairs, we additionally train Random and Reward variants at 20\% and 40\% under \textit{Mismatch-40\%}. 

Figure~\ref{fig:dpo_filter_rate_ablation} shows that increasing a prescribed filtering budget does not reproduce the training trajectory of DTV-Loo. Although more aggressive Reward-based Filtering improves over its lower-rate variants, DTV-Loo still exhibits faster optimization and achieves stronger overall performance, while increasing the Random Filtering rate provides no comparable benefit. These results indicate that the advantage of DTV-Loo cannot be attributed solely to a larger effective filtering ratio; rather, which preference pairs are removed is critical. More broadly, the sensitivity of Random and Reward-based Filtering to the prescribed removal rate highlights a limitation of fixed-rate filtering, as a single predefined ratio may not remain appropriate throughout training. In contrast, DTV and DTV-Loo use a fixed zero valuation threshold, allowing the effective filtering ratio to adapt dynamically to the evolving optimization state without tuning an explicit removal rate. This ablation tests sensitivity to the fixed filtering rate and it is not an evaluation of the theoretical DTV-$\lambda$ family.

\FloatBarrier

%%%%%%%%%%%%%%%%%%%%%%%%%%%%%%%%%%%%%%%%%%%%%%%%%%%%%%%%%%%%

% \newpage
% \input{checklist.tex}

\end{document}